\documentclass{article}
\usepackage[margin=1in]{geometry}
\usepackage{parskip}
\usepackage{needspace}

\usepackage[round]{natbib}

\usepackage[T1]{fontenc}    %
\usepackage{booktabs}       %

\usepackage{graphicx}
\usepackage[hypertexnames=false]{hyperref}%
\usepackage{color}
\usepackage{subcaption}
\usepackage{amsmath,amsthm,amssymb}
\usepackage{bm}
\usepackage{mathabx} %
\usepackage{mathtools}
\usepackage[most]{tcolorbox}
\hypersetup{colorlinks=true, linkcolor=blue, citecolor=blue}
\usepackage[nameinlink,capitalize,noabbrev]{cleveref} %
\usepackage{autonum} %

\theoremstyle{plain}
\newtheorem{theorem}{Theorem}%
\newtheorem{lemma}[theorem]{Lemma}
\newtheorem{corollary}[theorem]{Corollary}
\newtheorem{proposition}[theorem]{Proposition}

\newtheorem{condition}[theorem]{Condition}
\crefname{condition}{Condition}{Conditions}
\Crefname{condition}{Condition}{Conditions}

\theoremstyle{definition}
\newtheorem{definition}[theorem]{Definition}

\theoremstyle{remark}

\newtheorem{example}{Example}

\newcommand{\RR}{\mathbb{R}}

\newcommand{\NN}{\mathbb{N}}

\newcommand{\CC}{\mathbb{C}}
\newcommand{\EE}{\mathbb{E}}

\newcommand{\dd}{\mathrm{d}}

\renewcommand{\aa}{{\bm{a}}}

\newcommand{\xx}{{\bm{x}}}
\newcommand{\yy}{{\bm{y}}}
\newcommand{\zz}{{\bm{z}}}

\newcommand{\xxi}{{\bm{\xi}}}
\newcommand{\ttheta}{{\bm{\theta}}}

\newcommand{\Sch}{\mathcal{S}}
\newcommand{\G}{\mathcal{G}}
\newcommand{\F}{\mathcal{F}}
\renewcommand{\H}{\mathcal{H}}

\newcommand{\A}{\mathcal{A}}
\newcommand{\B}{\mathcal{B}}

\newcommand{\iprod}[1]{\langle#1\rangle}
\newcommand{\iiprod}[1]{(\!(#1)\!)}

\DeclareMathOperator{\sign}{sign}

\DeclareMathOperator{\supp}{supp}

\DeclareMathOperator{\im}{im}

\DeclareMathOperator{\erf}{erf}
\DeclareMathOperator{\pv}{pv}
\DeclareMathOperator{\fp}{fp}

\newcommand{\checks}[1]{\widecheck{#1}^\sharp}

\newtcolorbox{fouriermethodbox}{
  enhanced,
  breakable,
  colback=black!4,
  colframe=black!35,
  drop shadow=black!25,
  boxrule=0.5pt,
  arc=1mm,
  left=1.5mm,
  right=1.5mm,
  top=1.5mm,
  bottom=1.5mm,
  before skip=8pt,
  after skip=8pt
}

\allowdisplaybreaks

\title{Ghosts in Neural Networks: Existence, Structure and Role \\of Infinite-Dimensional Null Space}

\author{%
\textbf{Sho Sonoda}${}^{1,2}$ \hfill\small{\url{sho.sonoda@riken.jp}}\\
\textbf{Isao Ishikawa}${}^{3,1}$ \hfill\small{\url{ishikawa.isao.5s@kyoto-u.ac.jp}}\\
\textbf{Masahiro Ikeda}${}^{4,1}$ \hfill\small{\url{ikeda@ist.osaka-u.ac.jp}}\\
\small{${}^1$\textit{RIKEN AIP} }%
\small{${}^2$\textit{CyberAgent, Inc} }%
\small{${}^3$\textit{Kyoto University} }%
\small{${}^4$\textit{The University of Osaka}}\\
\phantom{\large{aaaaaaaaaaaaaaaaaaaaaaaaaaaaaaaaaaaaaaaaaaaaaaaaaaaaaaaaaaaaaaaaaaaaaaaa}}%
}

\begin{document}

\date{July 18th, 2026} %

\maketitle

\begin{abstract}
We study parameter nonuniqueness in continuous-width depth-two fully
connected neural networks.  Our main contribution is a direct method for
solving the neural-network equation $S[\gamma]=f$.  Starting from the
Fourier expression of the synthesis operator, separation of variables
produces a ridgelet particular solution and identifies every homogeneous
direction.  To isolate the argument, we first prove an abstract
reconstruction formula for unitary factorizations, yielding the adjoint,
normalized right inverse, and orthogonal solution geometry.  We then
specialize this formula to neural-network synthesis: for
tempered-distribution activations such as ReLU, we equip the activation
class $\A_{s,t}$ with a Hilbert structure, construct compatible coefficient
and parameter Hilbert spaces $\H_{s,t}$ and $\G_{s,t}$, and prove that
$S:\G_{s,t}\to L^2(\RR^m)$ is bounded.  The resulting ridgelet expansion
exhausts the null space and the complete solution set and identifies the
unique minimum-norm parameter distribution.  Concrete examples give
adjoint ridgelet functions for standard activations.  Further developments
show that finite-measure null elements admit normalized width-$N$
discretizations with $O(N^{-1/2})$ output error and characterize how
additive parameter perturbations can reveal information encoded in the null
space.  A Lean 4
blueprint for the main results is available at
\url{https://shosonoda.github.io/lean-ridgelet/}.

\end{abstract}

\section{Introduction}\label{sec:intro}
This section formulates parameter nonuniqueness as a neural-network
equation, presents the direct Fourier calculation that motivates the
analysis, and places the resulting solution method in context.

\subsection{Neural-Network Equation and Scope}

A scalar-output depth-two fully connected neural network with $N$ hidden
units has the form
\begin{align}
    f_N(\xx)
    =\sum_{j=1}^N c_j
      \sigma(\aa_j\cdot\xx-b_j).
    \label{eq:finite.network.intro}
\end{align}
Each map $\xx\mapsto\sigma(\aa_j\cdot\xx-b_j)$ is one hidden neuron,
and $c_j$ is its output weight.  Replacing the finite collection of neurons
by a continuously indexed family gives the integral representation
\begin{align}
    S[\gamma](\xx)
    := \int_{\RR^m\times\RR}
    \gamma(\aa,b)\sigma(\aa\cdot\xx-b)\dd\aa\dd b .
    \label{eq:conti.enn}
\end{align}
Thus $S[\gamma]$ is naturally interpreted as a continuous-width depth-two
network.  Conversely, a quadrature of \eqref{eq:conti.enn} has the form
\eqref{eq:finite.network.intro}; in the measure-valued formulation, a finite
network is represented by the atomic parameter measure
$\sum_{j=1}^N c_j\delta_{(\aa_j,b_j)}$.  Since Dirac measures do not belong
to the Hilbert parameter space used for the continuous theory, this precise
continuous-to-finite passage is treated separately in \cref{sec:finite}.

The distribution $\gamma$ records how the hidden parameters $(\aa,b)$ are
weighted.  Although the neuron is nonlinear in $(\aa,b)$, the map
$\gamma\mapsto S[\gamma]$ is linear.  This linearization makes it possible to
study neural-network parameters by operator theory.  It also exposes their
redundancy: the central problem of this paper is to solve the neural-network
equation
\begin{align}
    S[\gamma]=f
    \label{eq:neural.network.equation}
\end{align}
for a prescribed target $f$, including the complete homogeneous freedom
$\ker S$.  In particular, two parameter distributions represent the same
network function exactly when their difference belongs to $\ker S$.

The term ``ghost'' in the title follows the terminology of the classical
paper by \citet{Louis1981}, which first studied the detailed structure of the
null space of the Radon transform.  In the body of the paper we use the
standard terms \emph{null space} and \emph{null element}.  Our main scope is
the functional analysis of the continuous depth-two synthesis operator and
the complete solution of its neural-network equation.  Explicit adjoints,
finite-width approximation, numerical verification, and parameter
perturbations are treated afterward as consequences or independent tests of
the core theory.  Possible learning-theoretic implications are reserved for
discussion.

\subsection{Fourier expression method for the general solution}
\label{sec:fourier.solution.overview}

The main idea of this work is a direct method for discovering solutions of
\eqref{eq:neural.network.equation}.  Rather than postulating a ridgelet
transform, we begin with the Fourier expression of the network, separate its
two frequency variables, and then invert the resulting parameter formula.
The entire discovery calculation is displayed below.  Its analytic steps are
justified later in \cref{lem:fourier,thm:reconst,thm:bdd.S}.

\begin{fouriermethodbox}
\textbf{Fourier expression method.}
Fourier inversion in the preactivation variable gives, formally,
\begin{align}
    \sigma(\aa\cdot\xx-b)
    =\frac1{2\pi}\int_\RR
      \sigma^\sharp(\omega)
      e^{i\omega(\aa\cdot\xx-b)}\dd\omega.
\end{align}
Since $\gamma^\sharp$ denotes the Fourier transform of $\gamma$ in the bias
variable, substitution into the network integral yields
\begin{align}
    S[\gamma](\xx)
    &=\frac1{2\pi}\int_{\RR^m\times\RR}
      \gamma^\sharp(\aa,\omega)\sigma^\sharp(\omega)
      e^{i\omega\aa\cdot\xx}\dd\aa\dd\omega.
    \label{eq:fourier.network.intro}
\end{align}
Taking the Fourier transform in $\xx$ and using
$\int_{\RR^m}e^{-i\xx\cdot\zz}\dd\xx=(2\pi)^m\delta(\zz)$, we obtain
\begin{align}
    \widehat{S[\gamma]}(\xxi)
    &=\frac1{2\pi}\int_{\RR^m\times\RR}
      \gamma^\sharp(\aa,\omega)\sigma^\sharp(\omega)
      \left[\int_{\RR^m}
      e^{-i\xx\cdot(\xxi-\omega\aa)}\dd\xx\right]
      \dd\aa\dd\omega\notag\\
    &=(2\pi)^{m-1}\int_{\RR^m\times\RR}
      \gamma^\sharp(\aa,\omega)\sigma^\sharp(\omega)
      \delta(\xxi-\omega\aa)\dd\aa\dd\omega\notag\\
    &=(2\pi)^{m-1}\int_\RR
      \gamma^\sharp(\xxi/\omega,\omega)
      \sigma^\sharp(\omega)|\omega|^{-m}\dd\omega.
    \label{eq:fourier.solution.overview}
\end{align}
The last line separates the target frequency $\xxi$ from the
one-dimensional frequency $\omega$.  This suggests the
separation-of-variables choice
\begin{align}
    \gamma^\sharp(\xxi/\omega,\omega)
    :=\widehat f(\xxi)\overline{\rho^\sharp(\omega)}
    \label{eq:separation.ansatz.intro}
\end{align}
for a prescribed target $f$ and a compatible one-dimensional function
$\rho$.  Substitution gives
\begin{align}
    \widehat{S[\gamma]}(\xxi)
    &=(2\pi)^{m-1}\widehat f(\xxi)
      \int_\RR\sigma^\sharp(\omega)
      \overline{\rho^\sharp(\omega)}
      |\omega|^{-m}\dd\omega\notag\\
    &=\iiprod{\sigma,\rho}\widehat f(\xxi),
    \label{eq:pairing.solution.overview}\\
    \iiprod{\sigma,\rho}
    &:=(2\pi)^{m-1}\int_\RR
      \sigma^\sharp(\omega)\overline{\rho^\sharp(\omega)}
      |\omega|^{-m}\dd\omega.
    \label{eq:pairing.definition.intro}
\end{align}
Thus, whenever $\iiprod{\sigma,\rho}=1$, injectivity of the Fourier
transform gives
\begin{align}
    S[\gamma]=f.
    \label{eq:direct.particular.solution.intro}
\end{align}

It remains to identify the parameter distribution selected by
\eqref{eq:separation.ansatz.intro}.  Setting $\xxi=\omega\aa$ gives
\begin{align}
    \gamma^\sharp(\aa,\omega)
    =\widehat f(\omega\aa)\overline{\rho^\sharp(\omega)}.
\end{align}
Taking the inverse Fourier transform in the bias variable and expanding
$\widehat f$ then gives
\begin{align}
    \gamma(\aa,b)
    &=\frac1{2\pi}\int_\RR
      \widehat f(\omega\aa)\overline{\rho^\sharp(\omega)}
      e^{ib\omega}\dd\omega\notag\\
    &=\frac1{2\pi}\int_{\RR^m\times\RR}
      f(\xx)e^{-i\omega\aa\cdot\xx}
      \overline{\rho^\sharp(\omega)}
      e^{ib\omega}\dd\xx\dd\omega\notag\\
    &=\int_{\RR^m}f(\xx)
      \overline{\rho(\aa\cdot\xx-b)}\dd\xx
      =R[f;\rho](\aa,b).
    \label{eq:ridgelet.enn}
\end{align}
Consequently, the ridgelet transform is forced by the separated Fourier
expression rather than supplied as an external ansatz, and
\begin{align}
    S[R[f;\rho]]
    =\iiprod{\sigma,\rho}f.
    \label{eq:direct.reconstruction.intro}
\end{align}

The same calculation also reveals the homogeneous freedom.  Fix an
orthonormal basis $\{e_i\}_{i\in\NN}$ of $L^2(\RR^m)$ and compatible
ridgelet functions $\rho_0,\rho_1,\rho_2,\ldots$ such that
\begin{align}
    \iiprod{\sigma,\rho_0}=1,
    \qquad
    \iiprod{\sigma,\rho_j}=0\quad(j\ge1).
    \label{eq:direct.solution.conditions.intro}
\end{align}
For square-summable coefficients $(c_{ij})_{i,j\in\NN}$, consider the formal
series
\begin{align}
    \gamma
    :=R[f;\rho_0]
      +\sum_{i,j=1}^\infty c_{ij}R[e_i;\rho_j].
    \label{eq:direct.solution.template.intro}
\end{align}
Linearity, continuity and \eqref{eq:direct.reconstruction.intro} then give
\begin{align}
    S[\gamma]
    &=S[R[f;\rho_0]]
      +\sum_{i=1}^\infty\sum_{j=1}^\infty
       c_{ij}S[R[e_i;\rho_j]]\notag\\
    &=\iiprod{\sigma,\rho_0}f
      +\sum_{i=1}^\infty\sum_{j=1}^\infty
       c_{ij}\iiprod{\sigma,\rho_j}e_i
      =f.
    \label{eq:direct.general.solution.check}
\end{align}
\Cref{thm:struct} proves the converse: when the one-dimensional null
directions form a Hilbert basis, every solution is obtained in this way.
\end{fouriermethodbox}

The first term in \eqref{eq:direct.solution.template.intro} carries the
visible target, whereas the double series lies in the null space.  The
indices $i$ and $j$ resolve, respectively, the target-function direction
and the homogeneous one-dimensional direction.  The canonical choice
$\rho_0=c_\sigma^{-1}\sigma_*$ gives
$R[f;\rho_0]=S^\dagger[f]$, the unique minimum-norm particular solution.

Several analytic questions are hidden in the boxed discovery calculation.
For activations such as ReLU, $\sigma^\sharp$ is a tempered distribution, so
the weighted Fourier pairing is not an ordinary integral.  The operator $S$
is not bounded on all of $L^2(\RR^m\times\RR)$, and neither the exchange of
limits nor the infinite ridgelet series is automatic.  In particular,
boundedness is what permits a convergent sequence of parameters to be passed
through $S$.  It cannot be obtained by simply assuming
$\gamma\in L^2(\RR^m\times\RR)$: common activations are not integrable, and
the transformed parameter profile must act as the decay and regularity
factor in the distributional pairing.  We therefore construct the
activation, coefficient, and parameter Hilbert spaces $\A_{s,t}$,
$\H_{s,t}$, and $\G_{s,t}$.  Their Hilbert geometry also turns the apparent
arbitrariness of the parameters into an orthogonal decomposition, thereby
selecting a canonical minimum-norm solution.  Boundedness, the adjoint, and
the null-space expansion then provide the rigorous foundation for the direct
Fourier expression method above.

\subsection{Related Work and Context}

\subsubsection{Ridgelet Reconstruction}
The ridgelet transform was introduced in neural-network form by
\citet{Murata1996} and is closely related to the ridgelet and Radon-transform
constructions of \citet{Candes.PhD,Rubin.calderon}.  Reconstruction formulas
and admissibility conditions for non-integrable activations were developed
for depth-two networks by \citet{Sonoda2015acha}.  That work observed that
ridgelet functions orthogonal to the activation generate null elements.  The
present article goes further by deriving the ridgelet solution from the
Fourier expression of the neural-network equation and proving that the
resulting series exhausts the entire null space and hence the complete
solution set.

\subsubsection{Radon-Domain and Measure-Space Formulations}
More recent work has studied related, but distinct, forms of nonuniqueness
through Radon back-projection and variational function spaces.
\citet{Unser2023radon} constructed Radon-compatible Banach spaces, identified
the null space of an extended back-projection as a Banach complement, and
applied this framework to ReLU variational problems.  Banach-space
representer theorems likewise connect Radon-domain regularization with finite
shallow networks \citep{Parhi2021}.  Growth-controlled dual spaces and
Kantorovich--Rubinstein norms provide another framework for integral networks
on noncompact parameter domains \citep{Bartolucci2026}.  These measure- and
Banach-space theories complement our activation-adapted Hilbert-space
analysis of the synthesis operator itself.

\subsubsection{Finite Approximation}
Classical Maurey--Jones--Barron (MJB) arguments approximate an integral or
convex-hull representation by a width-$N$ network at the dimension-free rate
$O(N^{-1/2})$; see, for example,
\citet{Pisier1981,Jones1992,Barron1993,kainen.survey}.  We apply the same empirical
approximation mechanism to a measure-valued null representation.  The
conclusion is an approximate finite null relation on a bounded data domain,
not an exact identification of the null space of every finite
parametrization.  Exact finite relations arising from symmetries of the
activation are stated separately.

\subsubsection{Publication Context and Subsequent Work}
The Fourier expression method above first appeared in the 2021 preprint
version of this article \citep{Sonoda2021ghost}; that version contained the
original separation-of-variables solution method, the essential
$L^2$-boundedness argument, and the first null-space and general-solution
formulas.  Although these contributions remained unpublished, the method was
subsequently used and extended while its source was still a preprint.  Later
work developed a Fourier-slice method for closed-form ridgelet transforms
and surveyed ridgelet analysis \citep{Sonoda2024fourier}, while a
representation-theoretic method derived reconstruction principles for more
general equivariant feature maps \citep{Sonoda2025deep}.

The present article is therefore both the full proof of the original method
for the fully connected depth-two operator and a substantial extension of
the preprint.  New results include the abstract reconstruction formula in
\cref{prop:abstract.synthesis}, the orthogonal solution geometry of
\cref{thm:abstract.geometry}, explicit Hilbert structures for activations and
parameters, the normalized finite-approximation result in
\cref{thm:finite.null}, and the parameter-perturbation result
\cref{thm:perturbative.readout}.  The distinction between approximate and
exact finite null relations, and their numerical realization, is also made
explicit.

\subsection{Contributions and Organization}

The principal contribution is the direct Fourier expression method in
\cref{sec:fourier.solution.overview}: separation of the target and
one-dimensional frequencies produces a ridgelet solution and exposes its
homogeneous freedom.  \Cref{sec:abstract.solution} isolates the underlying
abstract reconstruction formula for unitary factorizations and the
orthogonal geometry of all solutions.
Sections~\ref{sec:pre}--\ref{sec:main} form the neural-network
specialization of this abstract formula.  \Cref{sec:pre} constructs the
activation, coefficient, and parameter Hilbert spaces and proves the
boundedness needed to justify limits.  \Cref{sec:fourier} realizes the
abstract reconstruction formula as ridgelet reconstruction from the Fourier
expression, and \cref{sec:main} identifies the adjoint, the entire null
space, the complete general solution, and its unique minimum-norm
representative.

After the core theory, \cref{sec:adjoint.examples} gives concrete examples
of adjoint ridgelet functions for ReLU, sigmoidal, and Gaussian activations.
Sections~\ref{sec:finite}--\ref{sec:readout} are further developments:
\cref{sec:finite,sec:numerical} connect continuous null elements with finite
networks analytically and numerically, while \cref{sec:readout} determines
how a parameter perturbation can make selected null-space information
visible.
\Cref{sec:discussion} states the scope of these conclusions.  Detailed
analytic proofs are collected in the appendices.  The main results are also
formalized in Lean 4; the blueprint is available at
\url{https://shosonoda.github.io/lean-ridgelet/}.

\section{Abstract Reconstruction Formula and Solution Geometry}
\label{sec:abstract.solution}

The calculation in \cref{sec:fourier.solution.overview} has two logically
distinct parts.  First, a coordinate transform separates the target variable
from a one-dimensional coefficient variable.  Second, the activation acts on
each coefficient vector through one bounded linear functional.  We isolate
this mechanism before introducing the Fourier transform, the weights, and the
activation-dependent spaces needed for the neural-network operator.

\subsection{Reconstruction from a Unitary Factorization}

The goal of this subsection is to identify the algebraic mechanism behind
ridgelet reconstruction without using any Fourier-specific notation.

\begin{definition}[Abstract synthesis and reconstruction operators]
Let $(X,\mu)$ be a $\sigma$-finite measure space, let $\H$ be a
separable complex Hilbert space, and set $\F:=L^2(X,\mu)$.
Let $\G$ be another Hilbert space and suppose that
\begin{align}
    T:\G\longrightarrow L^2(X,\mu;\H)
    \label{eq:abstract.unitary.T}
\end{align}
is unitary.  All inner products are linear in their first argument.  Given a
bounded linear functional $L\in\H^*$, define its pointwise lift by
$\widetilde L[u](x):=L[u(x)]$ for $u\in L^2(X,\mu;\H)$.
The pointwise bound for $L$ gives
$\|\widetilde L[u]\|_{\F}
\le\|L\|_{\H^*}\|u\|_{L^2(X;\H)}$, so this lift is a
bounded operator.  Define the synthesis operator
\begin{align}
    S:=\widetilde L T:\G\longrightarrow\F.
    \label{eq:abstract.factorization}
\end{align}
For $h\in\H$, also define the tensor embedding and its pullback to
parameter space by
\begin{align}
    J_h[f](x)&:=f(x)h,
    &
    R_h&:=T^*J_h.
    \label{eq:abstract.Rh}
\end{align}
Since $\|J_h[f]\|=\|h\|_{\H}\|f\|_{\F}$, both maps are
bounded.  Thus $R_h[f]=T^*[f\otimes h]$.  This is the abstract form of the
separated Fourier expression that later becomes a ridgelet transform.
\end{definition}

The following abstract reconstruction theorem is the first main result.  It
shows that every coefficient vector $h$ produces a reconstruction formula,
with normalization determined solely by the scalar $L[h]$.

\begin{theorem}[Reconstruction from a unitary factorization]
\label{prop:abstract.synthesis}
Let $h_L\in\H$ be the Riesz representer of $L$, so that
$L[h]=\iprod{h,h_L}_{\H}$, and put
$c_L:=\|h_L\|_{\H}^2=\|L\|_{\H^*}^2$.  Then, for every
$h\in\H$,
\begin{align}
    SR_h=L[h]I_{\F}.
    \label{eq:abstract.reconstruction}
\end{align}
Moreover,
\begin{align}
    S^*=R_{h_L},
    \qquad
    SS^*=c_LI_{\F},
    \qquad
    \|S^*[f]\|_{\G}^2=c_L\|f\|_{\F}^2.
    \label{eq:abstract.adjoint.identities}
\end{align}
\end{theorem}
\begin{proof}
See \cref{sec:proof.abstract.reconstruction}.  In brief,
$T[R_h[f]]=f\otimes h$ reduces reconstruction to the scalar identity
$L[f(x)h]=L[h]f(x)$, and unitarity transfers the Riesz identity for $L$
back to the parameter space.
\end{proof}

The identity $c_L=\|L\|_{\H^*}^2$ makes the normalization depend
only on the norm of the coefficient functional.  In the concrete theory this
functional is induced by the activation.

\subsection{Solution Geometry and Basis Expansion}

The next theorem turns the arbitrary part of a solution into orthogonal
geometry.  This is the reason for placing Hilbert structures on both the
coefficient and parameter spaces: without them one may still exhibit a right
inverse, but there is no canonical orthogonal projection or minimum-norm
representative.

\begin{theorem}[Orthogonal geometry of the solution set]
\label{thm:abstract.geometry}
Assume $L\ne0$ and define
\begin{align}
    S^\dagger:=c_L^{-1}S^*,
    \qquad
    P:=S^\dagger S.
    \label{eq:abstract.dagger.P}
\end{align}
Then
\begin{align}
    SS^\dagger&=I_{\F},
    &
    P^2&=P=P^*,
    &
    \ker P&=\ker S,
    &
    \im P&=\im S^*=(\ker S)^\perp,
    \label{eq:abstract.projection}
\end{align}
and
\begin{align}
    T[\ker S]=L^2(X,\mu;\ker L).
    \label{eq:abstract.kernel}
\end{align}
Consequently, for every $f\in\F$, the complete solution set is
\begin{align}
    \{\gamma\in\G:S[\gamma]=f\}
    =\{S^\dagger[f]+\eta:\eta\in\ker S\}.
    \label{eq:abstract.general.solution}
\end{align}
The element $S^\dagger[f]$ is the unique minimum-norm solution, and every
solution satisfies
\begin{align}
    \|\gamma\|_{\G}^2
    =\|S^\dagger[f]\|_{\G}^2
     +\|\gamma-S^\dagger[f]\|_{\G}^2.
    \label{eq:abstract.pythagorean}
\end{align}
\end{theorem}
\begin{proof}
See \cref{sec:proof.abstract.geometry}.  The identity $SS^*=c_LI$ makes
$P=c_L^{-1}S^*S$ an orthogonal projection, while the factorization
$S=\widetilde LT$ identifies its kernel pointwise with $\ker L$.
\end{proof}

In particular, $S^\dagger$ is the Moore--Penrose inverse of the surjective
operator $S$, and $P$ is its canonical orthogonal parameter projection.

The product-space formulation also makes the homogeneous freedom
constructive.  The following statement records the precise basis expansion
that underlies the double series in \eqref{eq:direct.solution.template.intro}.

\begin{theorem}[Hilbert-basis expansion in parameter space]
\label{thm:abstract.expansion}
Let $\{e_i\}_{i\in I}$ be an arbitrary Hilbert basis of $\F$.  For
$\gamma\in\G$, define the Bochner coefficients
\begin{align}
    h_i[\gamma]
    :=\int_X T[\gamma](x)\overline{e_i(x)}\dd\mu(x)
    \quad\text{in }\H.
    \label{eq:abstract.coefficients}
\end{align}
Then
\begin{align}
    \gamma=\sum_{i\in I}R_{h_i[\gamma]}[e_i]
    \quad\text{in }\G,
    \qquad
    \sum_{i\in I}\|h_i[\gamma]\|_{\H}^2
    =\|\gamma\|_{\G}^2.
    \label{eq:abstract.parameter.expansion}
\end{align}
The coefficients are unique.  Moreover,
\begin{align}
    \gamma\in\ker S
    \quad\Longleftrightarrow\quad
    h_i[\gamma]\in\ker L\quad\text{for every }i\in I.
    \label{eq:abstract.null.coefficients}
\end{align}
If $\{k_j\}_{j\in J}$ is a Hilbert basis of $\ker L$, every null
element has the equivalent double expansion
\begin{align}
    \gamma
    =\sum_{(i,j)\in I\times J}c_{ij}R_{k_j}[e_i],
    \qquad (c_{ij})\in\ell^2(I\times J).
    \label{eq:abstract.double.expansion}
\end{align}
\end{theorem}
\begin{proof}
See \cref{sec:proof.abstract.expansion}.  This is the Hilbert-basis
expansion of $T[\gamma]\in L^2(X;\H)$, transported back by $T^*$; Parseval
gives convergence and uniqueness.
\end{proof}

\section{Neural-Network Specialization: Hilbert Spaces and Boundedness}
\label{sec:pre}

We now realize the abstract scheme for the continuous-width network.  The
formal Fourier calculation requires a distributional pairing, a dilation,
and passage to limits.  The spaces below are chosen so that the Fourier
coordinate transform is unitary and the activation acts as a bounded
functional.  In particular, boundedness is not automatic: common
activations are not integrable, so the transformed parameter distribution
must supply the decay and regularity needed to pair with
$\sigma^\sharp$.  Throughout this concrete realization, $m\ge1$.

We deliberately retain the abstract notation and specialize it as
$X=\RR^m$, $\F=L^2(\RR^m)$, $\H=\H_{s,t}$,
$\G=\G_{s,t}$, and $L=L_\sigma$.  The Riesz data become
$h_L=h_\sigma$ and $c_L=c_\sigma$.  The operator symbols
$T,S,S^\dagger$, and $P$ therefore need no renaming, while the abstract
tensor solution $R_h[f]$ becomes the ridgelet transform $R[f;\rho]$ when
$h=h_\rho$.

\subsection{Fourier Conventions and the Activation Space}
\label{sec:activation.space}

The direct calculation uses Fourier transforms of activations that need not
be integrable.  We therefore fix the Fourier convention first and then
place the activation in a weighted Sobolev Hilbert space that accommodates
polynomial growth while retaining a continuous dual pairing.

For functions on $\RR^m$ and functions or tempered distributions on $\RR$,
respectively, we use
\begin{align}
    \widehat f(\xxi)&:=\int_{\RR^m}f(\xx)e^{-i\xx\cdot\xxi}\dd\xx,
    &
    \phi^\sharp(\omega)&:=\int_\RR\phi(b)e^{-ib\omega}\dd b.
\end{align}
The inverse transforms carry the factors $(2\pi)^{-m}$ and $(2\pi)^{-1}$.
Thus
\begin{align}
    \|\widehat f\|_{L^2(\RR^m)}^2
    =(2\pi)^m\|f\|_{L^2(\RR^m)}^2.
    \label{eq:plancherel.convention}
\end{align}

\begin{definition}[Activation space]
\label{def:activation.space}
Fix $s\in\RR$ and $t\ge0$, and write
$\iprod{\omega}:=(1+|\omega|^2)^{1/2}$.  The Bessel operator is defined by
$\iprod{\partial_\omega}^{t}[\phi^\sharp]
:=(\iprod{\cdot}^{t}\phi)^\sharp$.  The activation space is
\begin{align}
    \A_{s,t}
    :=\left\{\sigma\in\Sch'(\RR):
      \|\sigma\|_{\A_{s,t}}
      :=\big\|\iprod{\omega}^{s}
      \iprod{\partial_\omega}^{-t}[\sigma^\sharp]
      \big\|_{L^2(\RR)}<\infty\right\}.
    \label{eq:activation.space}
\end{align}
Equivalently, $\A_{s,t}=\iprod{\cdot}^{t}H^s(\RR)$, where
\begin{align}
    \iprod{\cdot}^{t}H^s(\RR)
    :=\{u\in\Sch'(\RR):\iprod{\cdot}^{-t}u\in H^s(\RR)\},
    \qquad
    \|u\|_{\iprod{\cdot}^{t}H^s}
    :=\|\iprod{\cdot}^{-t}u\|_{H^s}.
    \label{eq:weighted.sobolev.notation}
\end{align}
\end{definition}
This weighted Sobolev notation follows the microlocal-analysis convention of
\citet[Section~2.3]{Hintz2025microlocal}.  The Schwartz representation
theorem decomposes the tempered distributions into a union of weighted
Sobolev Hilbert spaces,
\begin{align}
    \Sch'(\RR)=\bigcup_{r,s\in\RR}\iprod{\cdot}^{r}H^s(\RR),
    \label{eq:schwartz.representation}
\end{align}
see \citet[Theorem~2.18]{Hintz2025microlocal}.  This result is the starting
point for selecting one fixed Hilbert member $\A_{s,t}$ rather than working
with the whole non-Hilbert space $\Sch'(\RR)$.  The index $t$ allows
polynomial growth, whereas $s$ measures Sobolev regularity after that growth
has been removed.

\begin{example}[Standard activations]
\label{ex:activation.membership}
The characterization
$\sigma\in\A_{s,t}\iff\sigma/\iprod{\cdot}^{t}\in H^s(\RR)$ gives concrete
choices.  The standard Gaussian
$(2\pi)^{-1/2}e^{-z^2/2}$ belongs to $\A_{s,t}$ for every $s\in\RR$ and
$t\ge0$.  The functions $\tanh z$ and the standard Gaussian cumulative
distribution function $\Phi_{\rm N}(z)$ belong to $\A_{s,t}$ for $s\le0$
and $t>1/2$.  The ReLU $z_+$ belongs to $\A_{s,t}$ for $s\le0$ and
$t>3/2$.  In particular, $\A_{0,2}$ contains all four activations.  The
unweighted space $L^2(\RR)$ would exclude the three nondecaying examples.
\end{example}

\begin{proposition}[Activation Hilbert structure]
\label{prop:activation.hilbert}
The space $\A_{s,t}$ is a Hilbert space, and
\begin{align}
    \sigma\longmapsto
    \iprod{\omega}^{s}\iprod{\partial_\omega}^{-t}[\sigma^\sharp]
    \label{eq:activation.isometry}
\end{align}
is an isometric isomorphism from $\A_{s,t}$ onto $L^2(\RR)$.
\end{proposition}
\begin{proof}
See \cref{sec:proof.activation.hilbert}.  The proof transports the Hilbert
structure of $L^2(\RR)$ through \eqref{eq:activation.isometry}.
\end{proof}

We assume henceforth that $0\ne\sigma\in\A_{s,t}$.

\subsection{Coefficient Hilbert Space and the Activation Functional}
\label{sec:fiber.space}

After the coordinate change below, a parameter distribution becomes a
section $\xx\mapsto T[\gamma](\xx,\cdot)$ of the trivial Hilbert bundle
$\RR^m\times\H_{s,t}\to\RR^m$.  The space $\H_{s,t}$ is the fiber; an
individual function $T[\gamma](\xx,\cdot)$ is a section value, which we call
a coefficient vector.  Equivalently, the section is an $\H_{s,t}$-valued
Bochner function, or an element of the Hilbert tensor product
$L^2(\RR^m)\widehat\otimes\H_{s,t}$.  These are compatible descriptions,
not different parameter models.

The coefficient norm must perform two jobs visible already in
\eqref{eq:fourier.solution.overview}.  The change
$\xxi=\omega\aa$ produces the Jacobian $|\omega|^m$, which accounts for the
first term below.  The pairing with the possibly distributional
$\sigma^\sharp$ requires weighted derivatives of the coefficient vector,
which accounts for the second.  Define
\begin{definition}[Coefficient space]
\label{def:coefficient.space}
\begin{align}
    \H_{s,t}
    &:=\overline{\Sch(\RR)}^{\|\cdot\|_{\H_{s,t}}},
    \label{eq:H.space}\\
    \|h\|_{\H_{s,t}}^2
    &:=(2\pi)^{m-1}\int_\RR|h(\omega)|^2|\omega|^m\dd\omega
      +\int_\RR
      \left|\iprod{\partial_\omega}^{t}[h](\omega)\right|^2
      \iprod{\omega}^{-2s}\dd\omega.
    \label{eq:H.norm}
\end{align}
\end{definition}
Neither term can generally be discarded.  Without the first, the inverse
coordinate transform need not return an $L^2$ parameter distribution.
Without the second, distributions such as the $\delta_0'$ term in the
Fourier transform of ReLU cannot act continuously on a coefficient vector.

The activation enters the abstract theory only through the action of
$\sigma^\sharp$.  We make this explicit in the following definition.
\begin{definition}[Activation functional]
\label{def:activation.functional}
\begin{align}
    L_\sigma[h]
    :=(2\pi)^{m-1}\int_\RR h(\omega)\sigma^\sharp(\omega)\dd\omega,
    \label{eq:ell.sigma}
\end{align}
The integral is initially understood by distributional duality on
$\Sch(\RR)$.
\end{definition}
Here
$\H_{s,t}^*$ denotes the continuous dual of $\H_{s,t}$, with its operator
norm.  Self-adjointness of the Bessel operator and Cauchy--Schwarz give
\begin{align}
    |L_\sigma[h]|
    &=(2\pi)^{m-1}\left|
      \int_\RR
      \bigl(\iprod{\omega}^{-s}
      \iprod{\partial_\omega}^{t}[h](\omega)\bigr)
      \bigl(\iprod{\omega}^{s}
      \iprod{\partial_\omega}^{-t}[\sigma^\sharp](\omega)\bigr)
      \dd\omega\right|\notag\\
    &\le(2\pi)^{m-1}
      \|h\|_{\H_{s,t}}\|\sigma\|_{\A_{s,t}}.
    \label{eq:fiber.pairing.bound}
\end{align}
Consequently, $L_\sigma$ extends uniquely to $\H_{s,t}$ and
\begin{align}
    \|L_\sigma\|_{\H_{s,t}^*}
    \le(2\pi)^{m-1}\|\sigma\|_{\A_{s,t}}.
    \label{eq:activation.to.dual.bound}
\end{align}
Thus the second term in the coefficient norm is precisely the convergence
factor that turns the formal activation pairing into the bounded functional
$L_\sigma$ required by the abstract theory.

\subsection{The Unitary Coordinate Transform and Parameter Graph Space}
\label{sec:parameter.space}

The Fourier expression suggests a change from parameter variables
$(\aa,b)$ to the input variable $\xx$ and the one-dimensional frequency
$\omega$.  The transform below performs this change; the graph norm then
retains exactly the additional coefficient regularity required by
$L_\sigma$.

\begin{definition}[Coordinate transform on Schwartz parameters]
\label{def:pointwise.coordinate.transform}
Set $\Theta:=\RR^m\times\RR$, and write
$F_\Theta[\gamma](\xxi,\omega)
:=\int_\Theta\gamma(\aa,b)e^{-i(\aa\cdot\xxi+b\omega)}\dd\aa\dd b$
for the full parameter-space Fourier transform.  For
$\gamma\in\Sch(\Theta)$, define
\begin{align}
    T_{\rm pt}[\gamma](\xx,\omega)
    &:=(2\pi)^{-m}F_\Theta[\gamma](-\omega\xx,\omega)\notag\\
    &=(2\pi)^{-m}\int_{\RR^m\times\RR}
      \gamma(\aa,b)e^{i\omega(\aa\cdot\xx-b)}\dd\aa\dd b
      =\checks{\gamma}(\omega\xx,\omega).
    \label{eq:T.transform}
\end{align}
\end{definition}
For each fixed $\xx$, this is a Schwartz function of $\omega$.  The formula
is a single integral transform whose phase is the neuron
preactivation $\aa\cdot\xx-b$.  Conceptually, it is the full Fourier
transform followed by pullback under
$(\xx,\omega)\mapsto(-\omega\xx,\omega)$; the mixed forward/inverse formula
in the last expression is only a convenient computational identity.

The pointwise formula need not define a Bochner $L^2$ coordinate for every
Schwartz parameter.  We therefore name its natural compatibility domain
$\Sch^T_{s,t}(\Theta):=\{\gamma\in\Sch(\Theta):
T_{\rm pt}[\gamma]\in L^2(\RR^m;\H_{s,t})\}$.
The low-frequency obstruction and the relation between this domain and the
Hilbert-space transform are detailed in \cref{sec:proof.T}.  In particular,
no inclusion $\Sch(\Theta)\subset\G_{s,t}$ is being asserted.

The following elementary measure identity isolates a point that is easy to
lose in a formal dilation calculation.

\begin{lemma}[Weighted dilation identity]
\label{lem:weighted.dilation}
If $g:\RR^m\times\RR\to[0,\infty)$ is Borel measurable, then
\begin{align}
    \int_{\RR^m\times\RR}g(\aa,\omega)\dd\aa\dd\omega
    =\int_{\RR^m\times\RR}
      g(-\omega\xx,\omega)|\omega|^m\dd\xx\dd\omega,
    \label{eq:weighted.dilation}
\end{align}
with equality also when the common value is infinite.
\end{lemma}
\begin{proof}
See \cref{sec:proof.weighted.dilation}.  The ordinary change of variables
applies on every slice $\omega\ne0$, and the exceptional slice is null.
\end{proof}

For $\gamma\in\Sch(\Theta)$, Plancherel on the full
$(m+1)$-dimensional parameter space and
\cref{lem:weighted.dilation} yield
\begin{align}
    \|\gamma\|_{L^2(\RR^m\times\RR)}^2
    =(2\pi)^{m-1}\int_{\RR^m\times\RR}
      |T_{\rm pt}[\gamma](\xx,\omega)|^2|\omega|^m\dd\xx\dd\omega.
    \label{eq:T.L2}
\end{align}
Consequently, $T_{\rm pt}$ extends in the weighted $L^2$ norm to a unitary
map $T_0:L^2(\Theta)\to\mathcal W_m$, where the weighted
coordinate space $\mathcal W_m$ is defined in
\eqref{eq:weighted.coordinate.space}.  Its construction and inverse are
given in \cref{sec:proof.T}.  The final unitary coordinate transform is the
restriction of $T_0$ to the graph domain below.

The remaining term needed for the activation pairing is generally not
controlled by this $L^2$ norm.  We therefore use the graph space
\begin{definition}[Parameter space and unitary coordinate transform]
\label{def:parameter.space}
\begin{align}
    \G_{s,t}
    :=\{\gamma\in L^2(\RR^m\times\RR):
       T_0[\gamma]\in L^2(\RR^m;\H_{s,t})\},
    \label{eq:G.space}
\end{align}
where membership means that the weighted coordinate $T_0[\gamma]$ admits
the indicated $\H_{s,t}$-valued representative.  For
$\gamma\in\G_{s,t}$, write $T[\gamma]:=T_0[\gamma]$ and call $T$ the
\emph{unitary coordinate transform}.  Its role is to place the synthesis
problem in the product coordinates of \cref{sec:abstract.solution}.
Equip $\G_{s,t}$ with
\begin{align}
    \|\gamma\|_{\G_{s,t}}^2
    :=\|\gamma\|_{L^2(\RR^m\times\RR)}^2
      +\int_{\RR^m\times\RR}
      \left|\iprod{\partial_\omega}^{t}
      [T[\gamma](\xx,\omega)]\right|^2
      \iprod{\omega}^{-2s}\dd\xx\dd\omega.
    \label{eq:G.norm}
\end{align}
\end{definition}
The base term retains the original parameter distribution; the graph term
is exactly the coefficient regularity used in
\eqref{eq:fiber.pairing.bound}.  Omitting the graph term would make the ReLU
pairing unbounded, whereas omitting the base term would lose control of the
underlying $L^2$ parameter.

For the inverse integral formula, use the coordinate test class
$\mathcal E_{s,t}:=\Sch(\RR^m)\otimes_{\mathrm{alg}}\Sch(\RR)
\subset L^2(\RR^m;\H_{s,t})$,
where a finite tensor sum is regarded as an $\H_{s,t}$-valued function.
This class is dense in the displayed Bochner space; see
\cref{sec:proof.T}.

\begin{proposition}[Concrete unitary coordinate transform]
\label{prop:T.unitary}
The graph-domain transform is a unitary map
\begin{align}
    T:\G_{s,t}\longrightarrow L^2(\RR^m;\H_{s,t}).
    \label{eq:G.Bochner}
\end{align}
For every $u\in\mathcal E_{s,t}$, its adjoint, equivalently its inverse, is
\begin{align}
    T^*[u](\aa,b)
    =\frac1{2\pi}\int_{\RR^m\times\RR}
      u(\xx,\omega)e^{-i\omega(\aa\cdot\xx-b)}
      |\omega|^m\dd\xx\dd\omega.
    \label{eq:T.adjoint.inverse}
\end{align}
\end{proposition}
\begin{proof}
See \cref{sec:proof.T}.  Identity \eqref{eq:T.L2} identifies the first term
of the Bochner norm with the $L^2$ parameter norm; the second term is the
graph term by definition.  The proof also constructs the inverse, proves
surjectivity, and establishes the integral formula first on
$\mathcal E_{s,t}$.
\end{proof}

On the compatibility domain $\Sch^T_{s,t}(\Theta)$, the Hilbert-space
transform $T[\gamma]$ is represented almost everywhere by the pointwise
integral $T_{\rm pt}[\gamma]$ in \eqref{eq:T.transform}; this follows from
the construction in \cref{sec:proof.T}.  Thus \cref{prop:T.unitary}
realizes the abstract unitary map
$T:\G\to L^2(X;\H)$ as
$T:\G_{s,t}\to L^2(\RR^m;\H_{s,t})$.

\subsection{Factorization and Boundedness of the Neural-Network Operator}

We now combine the unitary transform with the bounded activation functional.
This factorization is the bridge between the original integral and the
abstract synthesis operator, and its pointwise estimate supplies the
boundedness needed to pass limits through $S$.

\begin{definition}[Hilbert-space synthesis operator]
\label{def:hilbert.synthesis}
For Schwartz data, the classical integral representation is
\begin{align}
    S[\gamma](\xx)
    :=\int_{\RR^m\times\RR}
      \gamma(\aa,b)\sigma(\aa\cdot\xx-b)\dd\aa\dd b.
    \label{eq:intrep}
\end{align}
Fourier inversion in the preactivation shows that, on the completed Hilbert
spaces, its natural definition is the factorization
\begin{align}
    S&=\widetilde L_\sigma T,
    &
    S[\gamma](\xx)
    &=L_\sigma[T[\gamma](\xx,\cdot)]
      =(2\pi)^{m-1}\int_\RR
      T[\gamma](\xx,\omega)\sigma^\sharp(\omega)\dd\omega.
    \label{eq:S.T}
\end{align}
This is exactly the abstract factorization \eqref{eq:abstract.factorization}
under the specialization above.
\end{definition}

\begin{theorem}[Boundedness of $S$]
\label{thm:bdd.S}
For every $\sigma\in\A_{s,t}$, \eqref{eq:S.T} defines a bounded operator
$S:\G_{s,t}\to L^2(\RR^m)$ satisfying
\begin{align}
    \|S[\gamma]\|_{L^2(\RR^m)}
    \le(2\pi)^{m-1}\|\sigma\|_{\A_{s,t}}
       \|\gamma\|_{\G_{s,t}}.
    \label{eq:S.bound}
\end{align}
It agrees with \eqref{eq:intrep} whenever the classical integral is
defined.
\end{theorem}
\begin{proof}
Apply the pointwise estimate \eqref{eq:fiber.pairing.bound}, integrate in
$\xx$, and use the unitary identity \eqref{eq:G.Bochner}.  The derivation of
\eqref{eq:S.T}, the density argument, and agreement with the classical
integral are given in \cref{sec:proof.bddS}.
\end{proof}

This boundedness is what permits limits of parameter distributions and
infinite ridgelet expansions to pass through $S$.  It supplies the analytic
step that the formal solution in the Introduction cannot provide by itself.

\section{Neural-Network Specialization: Ridgelet Reconstruction from the Fourier Expression}
\label{sec:fourier}

We next translate the abstract tensors $T^*[f\otimes h]$ back into the
standard ridgelet notation.  The purpose is twofold: to justify the
Fourier calculation in the Introduction and to identify precisely which
one-dimensional ridgelet functions give elements of the parameter Hilbert
space.  The key idea is to absorb the Jacobian $|\omega|^{-m}$ into the
coefficient vector $h_\rho$.

\subsection{Ridgelet Transform and Fourier Formulas}

\begin{definition}[Ridgelet transform and activation pairing]
\label{def:ridgelet.transform}
On a test class, the classical ridgelet transform is
\begin{align}
    R[f;\rho](\aa,b)
    :=\int_{\RR^m}f(\xx)
      \overline{\rho(\aa\cdot\xx-b)}\dd\xx,
    \label{eq:ridgelet.definition}
\end{align}
The associated activation pairing and weighted norm are
\begin{align}
    \iiprod{\sigma,\rho}
    &:=(2\pi)^{m-1}\int_\RR
      \sigma^\sharp(\omega)\overline{\rho^\sharp(\omega)}
      |\omega|^{-m}\dd\omega,
    \label{eq:admissibility.pairing}\\
    \|\rho\|_{L_m^2}^2
    &:=(2\pi)^{m-1}\int_\RR
      |\rho^\sharp(\omega)|^2|\omega|^{-m}\dd\omega.
    \label{eq:Lm.norm}
\end{align}
The pairing is understood by the continuous duality above whenever it is not
an ordinary integral.  Absorb the dilation weight into
\begin{align}
    h_\rho(\omega)
    :=|\omega|^{-m}\overline{\rho^\sharp(\omega)},
    \qquad
    \B_{s,t}:=\{\rho\in\Sch'(\RR):h_\rho\in\H_{s,t}\}.
    \label{eq:compatible.ridgelets}
\end{align}
Then $\iiprod{\sigma,\rho}=L_\sigma[h_\rho]$ and
$\B_{s,t}\subset L_m^2$.  For general $f\in L^2(\RR^m)$ and
$\rho\in\B_{s,t}$, the rigorous Hilbert-space definition is
\begin{align}
    R[f;\rho]:=T^*[f\otimes h_\rho]\in\G_{s,t}.
    \label{eq:ridgelet.Hilbert.definition}
\end{align}
\end{definition}
The next lemma shows that this extension agrees with
\eqref{eq:ridgelet.definition} whenever the classical integral exists.

\begin{lemma}[Fourier expressions for $S$ and $R$]
\label{lem:fourier}
For $f\in L^2(\RR^m)$, $\rho\in\B_{s,t}$, and
$\gamma\in\G_{s,t}$,
\begin{align}
    R[f;\rho]^\sharp(\aa,\omega)
    &=\widehat f(\omega\aa)\overline{\rho^\sharp(\omega)},
    \label{eq:fourierR}\\
    \widehat{S[\gamma]}(\xxi)
    &=(2\pi)^{m-1}\int_\RR
      \gamma^\sharp(\xxi/\omega,\omega)
      \sigma^\sharp(\omega)|\omega|^{-m}\dd\omega.
    \label{eq:fourierS}
\end{align}
Both identities hold in the corresponding $L^2$ spaces, with the second
integral interpreted through \eqref{eq:fiber.pairing.bound}.  Moreover,
\begin{align}
    T[R[f;\rho]]&=f\otimes h_\rho,
    &
    R[f;\rho]&=T^*[f\otimes h_\rho],
    &
    \|R[f;\rho]\|_{\G_{s,t}}
      &=\|f\|_{L^2}\|h_\rho\|_{\H_{s,t}}.
    \label{eq:R.T}
\end{align}
\end{lemma}
\begin{proof}
See \cref{sec:proof.fourier}.  The formulas are first computed on Schwartz
functions and then extended by \cref{prop:T.unitary,thm:bdd.S}.
\end{proof}

\subsection{Reconstruction and Direct Solutions}
\label{sec:direct.fourier.solution}

Thus the ridgelet transform is precisely the abstract separated solution
$R_{h_\rho}[f]$ from \eqref{eq:abstract.Rh}.  The following reconstruction
formula is consequently a concrete instance of
\eqref{eq:abstract.reconstruction}.

\begin{theorem}[Reconstruction formula]
\label{thm:reconst}
If $f\in L^2(\RR^m)$, $\sigma\in\A_{s,t}$, and
$\rho\in\B_{s,t}$, then
\begin{align}
    S[R[f;\rho]]=\iiprod{\sigma,\rho}f
    \quad\text{in }L^2(\RR^m).
    \label{eq:reconstruction}
\end{align}
\end{theorem}
\begin{proof}
By \eqref{eq:R.T}, the coefficient section is $f\otimes h_\rho$.
Equations \eqref{eq:S.T} and
$L_\sigma[h_\rho]=\iiprod{\sigma,\rho}$ give the result.  A direct Fourier
calculation is also recorded in \cref{sec:proof.fourier}.
\end{proof}

Choose $\rho_0\in\B_{s,t}$ with
$\iiprod{\sigma,\rho_0}=1$; the canonical choice is constructed below.  For
$f\in L^2(\RR^m)$, the separated Fourier expression
\begin{align}
    \gamma_f^\sharp(\aa,\omega)
    :=\widehat f(\omega\aa)\overline{\rho_0^\sharp(\omega)}
    \label{eq:direct.particular.fourier}
\end{align}
is $\gamma_f=R[f;\rho_0]$, and
\cref{thm:reconst} gives $S[\gamma_f]=f$.  Hence every solution has the
form
\begin{align}
    \gamma=R[f;\rho_0]+\eta,
    \label{eq:direct.general.solution}
\end{align}
where $\eta\in\G_{s,t}$ satisfies, for almost every $\xxi$,
\begin{align}
    (2\pi)^{m-1}\int_\RR
      \eta^\sharp(\xxi/\omega,\omega)
      \sigma^\sharp(\omega)|\omega|^{-m}\dd\omega=0.
    \label{eq:direct.homogeneous.fourier}
\end{align}
Conversely, every such $\eta$ may be added without changing the represented
function.  The abstract expansion theorem proves completeness of this
description; the next section states its concrete ridgelet form.

\section{Neural-Network Specialization: Adjoint, Null Space, and Complete General Solution}
\label{sec:main}

The reconstruction theorem supplies particular solutions, but the complete
solution requires the adjoint range and the whole null space.  We obtain
both from the Riesz representer of the activation functional.  The abstract
orthogonal decomposition then gives the canonical solution, while the
Hilbert-basis expansion converts every null component into ridgelet terms.

\subsection{Adjoint and Canonical Solution}

\begin{definition}[Adjoint ridgelet function]
\label{def:adjoint.ridgelet}
The concrete counterpart of the abstract Riesz vector $h_L$ is the Riesz
representer $h_\sigma\in\H_{s,t}$ of $L_\sigma$:
\begin{align}
    L_\sigma[h]=\iprod{h,h_\sigma}_{\H_{s,t}}.
    \label{eq:h.sigma}
\end{align}
Equivalently, it is the unique weak solution of
\begin{align}
    (2\pi)^{m-1}\int_\RR h(\omega)\sigma^\sharp(\omega)\dd\omega
    &=(2\pi)^{m-1}\int_\RR
      h(\omega)\overline{h_\sigma(\omega)}|\omega|^m\dd\omega\notag\\
    &\quad+\int_\RR
      \iprod{\partial_\omega}^{t}[h](\omega)
      \overline{\iprod{\partial_\omega}^{t}[h_\sigma](\omega)}
      \iprod{\omega}^{-2s}\dd\omega.
    \label{eq:h.sigma.weak}
\end{align}
Define
\begin{align}
    \sigma_*^\sharp(\omega)
    :=|\omega|^m\overline{h_\sigma(\omega)},
    \qquad
    c_\sigma:=\|h_\sigma\|_{\H_{s,t}}^2
      =\|L_\sigma\|_{\H_{s,t}^*}^2>0.
    \label{eq:sigma.star}
\end{align}
The function $\sigma_*$ is called the adjoint ridgelet function.
\end{definition}
Here $h_{\sigma_*}=h_\sigma$, so $\sigma_*\in\B_{s,t}$.  Moreover,
$c_\sigma>0$: otherwise $L_\sigma$ would vanish on the dense subspace
$\Sch(\RR)\subset\H_{s,t}$,
which would imply $\sigma^\sharp=0$ and contradict $\sigma\ne0$.

\begin{lemma}[Concrete adjoint and canonical ridgelet]
\label{lem:adj.S}
The Hilbert adjoint is
\begin{align}
    S^*[f]=R[f;\sigma_*].
    \label{eq:adjoint.ridgelet}
\end{align}
It satisfies
\begin{align}
    SS^*&=c_\sigma I_{L^2(\RR^m)},
    &
    \|S^*[f]\|_{\G_{s,t}}^2
      &=c_\sigma\|f\|_{L^2}^2,
    &
    \iiprod{\sigma,\sigma_*}&=c_\sigma.
    \label{eq:adjoint.identities}
\end{align}
\end{lemma}
\begin{proof}
Since $h_{\sigma_*}=h_\sigma$, equation \eqref{eq:R.T} identifies
$R[f;\sigma_*]$ with the abstract operator $R_{h_\sigma}[f]$.  The claims
therefore follow from \cref{prop:abstract.synthesis}.  The inner-product
calculation is written out in \cref{sec:proof.adjS}.
\end{proof}

\begin{definition}[Canonical solution and parameter projection]
\label{def:canonical.solution}
The normalized right inverse and orthogonal parameter projection are
\begin{align}
    S^\dagger:=c_\sigma^{-1}S^*,
    \qquad
    P:=S^\dagger S=c_\sigma^{-1}S^*S.
    \label{eq:dagger.and.P}
\end{align}
\end{definition}
By \cref{thm:abstract.geometry},
\begin{align}
    SS^\dagger&=I,
    &
    P^2&=P=P^*,
    &
    \im P&=\im S^*=(\ker S)^\perp,
    &
    \ker P&=\ker S.
    \label{eq:projection.properties}
\end{align}
The normalization by $c_\sigma$ is necessary unless $c_\sigma=1$.

\subsection{Null Space and General Solution}

It remains to express the orthogonal null component in ridgelet variables.
The basis expansion below proves that the Fourier construction is exhaustive,
not merely a source of particular null elements.

\begin{theorem}[Ridgelet characterization of the null space and general solution]
\label{thm:struct}
Let $\{e_i\}_{i\in\NN}$ be an orthonormal basis of $L^2(\RR^m)$.  Then:
\begin{enumerate}
    \item
    \begin{align}
        T[\ker S]
        =L^2(\RR^m;\ker L_\sigma),
        \label{eq:null.tensor.characterization}
    \end{align}
    equivalently, $\gamma_0\in\ker S$ if and only if
    \begin{align}
        L_\sigma[T[\gamma_0](\xx,\cdot)]=0
        \quad\text{for almost every }\xx.
        \label{eq:null.fiber.condition}
    \end{align}

    \item Every $\gamma_0\in\ker S$ has the unique expansion
    \begin{align}
        \gamma_0=\sum_{i=1}^\infty R[e_i;\rho_i]
        \quad\text{in }\G_{s,t},
        \qquad
        \iiprod{\sigma,\rho_i}=0,
        \label{eq:null.ridgelet.series}
    \end{align}
    where
    \begin{align}
        h_i
        &:=\int_{\RR^m}T[\gamma_0](\xx,\cdot)
          \overline{e_i(\xx)}\dd\xx
          \quad\text{in }\H_{s,t},
        \label{eq:hi.coefficient}\\
        \rho_i^\sharp(\omega)
        &:=|\omega|^m\overline{h_i(\omega)}.
        \label{eq:rhoi.coefficient}
    \end{align}

    \item For every $f\in L^2(\RR^m)$ and every
    $\rho_0\in\B_{s,t}$ with $\iiprod{\sigma,\rho_0}=1$, a parameter
    $\gamma\in\G_{s,t}$ solves $S[\gamma]=f$ if and only if it has the
    unique representation
    \begin{align}
        \gamma
        =R[f;\rho_0]+\sum_{i=1}^\infty R[e_i;\rho_i]
        \quad\text{in }\G_{s,t},
        \qquad
        \iiprod{\sigma,\rho_i}=0,
        \label{eq:general.solution.series}
    \end{align}
    where the coefficients $\rho_i$ are obtained from the null residual
    $\gamma-R[f;\rho_0]$ by \eqref{eq:hi.coefficient} and
    \eqref{eq:rhoi.coefficient}.  Equivalently, the complete solution set is
    \begin{align}
        \{R[f;\rho_0]+\gamma_0:\gamma_0\in\ker S\}
        =\{S^\dagger[f]+\gamma_0:\gamma_0\in\ker S\}.
        \label{eq:general.solution}
    \end{align}
\end{enumerate}
\end{theorem}
\begin{proof}
Apply \cref{thm:abstract.geometry,thm:abstract.expansion} with
$\G=\G_{s,t}$, $\H=\H_{s,t}$, and $L=L_\sigma$.
Equation \eqref{eq:R.T} converts
each abstract term $R_{h_i}[e_i]$ into $R[e_i;\rho_i]$ through
\eqref{eq:rhoi.coefficient}.  The detailed coefficient translation and the
connection with the Fourier expression method are given in
\cref{sec:proof.struct}.
\end{proof}

The minimum-norm and Pythagorean properties of the canonical representative
$S^\dagger[f]$ are already supplied by \cref{thm:abstract.geometry}.
\Cref{thm:struct} records the additional concrete content: the complete
Fourier/ridgelet parametrization of the null residual and hence of every
solution.

The single-index series bundles the two independent tensor directions.  If
$\{h_{\rho_j}\}_{j\in\NN}$ is an orthonormal basis of $\ker L_\sigma$, then
every null element also has a unique square-summable double expansion
\begin{align}
    T[\gamma_0]
    =\sum_{i,j=1}^\infty c_{ij}e_i\otimes h_{\rho_j},
    \qquad
    \gamma_0
    =\sum_{i,j=1}^\infty c_{ij}R[e_i;\rho_j].
    \label{eq:concrete.double.expansion}
\end{align}
This is the rigorous version of
\eqref{eq:direct.solution.template.intro}.  In particular,
\begin{align}
    \ker S\cong L^2(\RR^m)\widehat\otimes\ker L_\sigma,
    \label{eq:null.tensor.product}
\end{align}
and every parameter has the orthogonal decomposition
\begin{align}
    \gamma=P[\gamma]+(I-P)[\gamma],
    \qquad
    P[\gamma]\in(\ker S)^\perp,
    \quad
    (I-P)[\gamma]\in\ker S.
    \label{eq:orthogonal.decomposition}
\end{align}
Thus the arbitrary part of a parameter distribution is exactly its null
component, while $P[\gamma]=S^\dagger[S[\gamma]]$ is determined by the
represented function.

\section{Examples of Adjoint Ridgelet Functions: ReLU, Sigmoidal, and Gaussian Activations}
\label{sec:adjoint.examples}

The abstract theory identifies the canonical ridgelet function $\sigma_*$
through the Riesz representer of $L_\sigma$, but this characterization leaves
its concrete form implicit.  Since $\sigma_*$ determines both the adjoint
$S^*$ and the minimum-norm solution operator $S^\dagger$, it is natural to
ask what this function looks like for familiar activations.  We answer this
question by rewriting the weak Riesz equation as a one-dimensional
resolvent problem in the Fourier variable and evaluating it for ReLU,
$\tanh$, the Gaussian cumulative distribution function, and the Gaussian
density.

The ridgelet function $\sigma_*$ depends not only on the activation
$\sigma$ but also on the Hilbert geometry specified by $(s,t)$.  The weak
equation \eqref{eq:h.sigma.weak} gives a useful concrete Fourier
description.  For brevity, put $C_m:=(2\pi)^{m-1}$ and
$r_\sigma:=\overline{h_\sigma}$.
Then, in the weak sense,
\begin{align}
    \left[
      C_m|\omega|^m
      +\iprod{\partial_\omega}^{t}
       \iprod{\omega}^{-2s}
       \iprod{\partial_\omega}^{t}
    \right]r_\sigma
    &=C_m\sigma^\sharp,
    \qquad
    \sigma_*^\sharp=|\omega|^m r_\sigma.
    \label{eq:adjoint.riesz.operator}
\end{align}
Here the expression in brackets denotes composition of the multiplication
and Bessel operators in the displayed order.

\begin{example}[Standard activations]
For a common setting that contains all the examples below, take
$(s,t)=(0,2)$ and define the positive one-dimensional operator
\begin{align}
    K_m:=(1-\partial_\omega^2)^2+C_m|\omega|^m.
    \label{eq:common.adjoint.operator}
\end{align}
The four activations
\begin{align}
    \sigma_{\rm R}(z)&:=z_+,
    &
    \sigma_{\rm T}(z)&:=\tanh z,\notag\\
    \sigma_{\rm C}(z)&:=\Phi_{\rm N}(z)
      :=\frac{1}{\sqrt{2\pi}}\int_{-\infty}^z e^{-u^2/2}\dd u
      =\frac12\left(1+\erf\frac{z}{\sqrt2}\right),
    &
    \sigma_{\rm G}(z)&:=\frac{1}{\sqrt{2\pi}}e^{-z^2/2}
    \label{eq:example.activations}
\end{align}
belong to $\A_{0,2}$.  With the Fourier convention of
\cref{sec:pre}, their transforms are
\begin{align}
    \sigma_{\rm R}^\sharp
    &=-\fp\frac{1}{\omega^2}+i\pi\delta_0',
    &
    \sigma_{\rm T}^\sharp
    &=-i\pi\,\pv\frac{1}{\sinh(\pi\omega/2)},\notag\\
    \sigma_{\rm C}^\sharp
    &=\pi\delta_0
      -i e^{-\omega^2/2}\pv\frac{1}{\omega},
    &
    \sigma_{\rm G}^\sharp
    &=e^{-\omega^2/2}.
    \label{eq:activation.fourier.examples}
\end{align}
The finite-part and Dirac terms in the ReLU formula follow from the
classical table of Fourier transforms of generalized functions
\citep[pp.~359--360, Entries~22--23]{GelfandShilov}.

Substitution into \eqref{eq:adjoint.riesz.operator} gives the corresponding
adjoint ridgelet functions:
\begin{align}
    (\sigma_{\rm R})_*^\sharp
    &=C_m|\omega|^mK_m^{-1}
      \left[-\fp\frac{1}{\omega^2}+i\pi\delta_0'\right],
    \notag\\
    (\sigma_{\rm T})_*^\sharp
    &=-i\pi C_m|\omega|^mK_m^{-1}
      \left[\pv\frac{1}{\sinh(\pi\omega/2)}\right],
    \notag\\
    (\sigma_{\rm C})_*^\sharp
    &=C_m|\omega|^mK_m^{-1}
      \left[\pi\delta_0
      -i e^{-\omega^2/2}\pv\frac{1}{\omega}\right],
    \notag\\
    (\sigma_{\rm G})_*^\sharp
    &=C_m|\omega|^mK_m^{-1}[e^{-\omega^2/2}].
    \label{eq:adjoint.ridgelet.examples}
\end{align}
The inverse $K_m^{-1}$ is understood in the weak sense supplied by the
Riesz problem; thus these are concrete one-dimensional resolvent formulas.
If either sigmoidal activation is considered separately, the smaller choice
$(s,t)=(0,1)$ is possible, with $K_m$ replaced by
$1-\partial_\omega^2+C_m|\omega|^m$.

For the standard Gaussian alone, one may take $(s,t)=(0,0)$.  In this case
the resolvent becomes a Fourier multiplier and yields the closed expression
\begin{align}
    (\sigma_{\rm G})_*^\sharp(\omega)
    =\frac{C_m|\omega|^m}{1+C_m|\omega|^m}
      e^{-\omega^2/2}.
    \label{eq:gaussian.adjoint.closed}
\end{align}
\end{example}
The derivation of these formulas is recorded in
\cref{sec:proof.adjoint.examples}.

\section{Further Developments: Finite-Width Approximation of Null Elements}\label{sec:finite}

The null space above was obtained for a continuous-width network, so one
may wonder whether it is merely a pathology of the infinite-width
description.  This section explains how the same parameter redundancy
leaves a finite-width trace.  The key observation is that synthesis is an
integral: a nonzero null parameter measure can therefore be discretized by
sampling or quadrature into atomic measures, hence into finite networks.
The resulting normalized finite sums form an approximate null sequence,
with an explicit $N^{-1/2}$ output bound.  Exact finite relations caused by
algebraic symmetries are treated separately, since a fixed finite feature
matrix need not have a nontrivial kernel.

\subsection{Measure-Valued Synthesis}

The Hilbert space $\G_{s,t}$ does not contain Dirac measures, whereas a finite
network is naturally represented by a finite atomic measure.  We therefore
use a measure-valued formulation in this section.  We also localize the
output norm: an individual ridge function usually does not belong to
$L^2(\RR^m)$, although it is square-integrable on a bounded data domain.

\begin{definition}[Measure-valued synthesis]
\label{def:measure.synthesis}
Let $\nu$ be a Borel probability measure on $\RR^m$ and put
$\H_\nu:=L^2(\RR^m,\nu)$.  Recall that
$\Theta=\RR^m\times\RR$.  In this section, assume additionally that
$\sigma$ is a Borel measurable function and set
\begin{align}
    \Phi_{(\aa,b)}(\xx):=\sigma(\aa\cdot\xx-b).
    \label{eq:finite.feature}
\end{align}
For a finite complex Radon measure $\mu$ on $\Theta$, let $|\mu|$ denote
its total-variation measure.  Whenever
\begin{align}
    E_\nu(\mu)
    :=\int_\Theta
      \|\Phi_\ttheta\|_{\H_\nu}^2\dd|\mu|(\ttheta)
    <\infty,
    \label{eq:feature.energy}
\end{align}
the Bochner integral
\begin{align}
    S_\nu[\mu]
    :=\int_\Theta\Phi_\ttheta\dd\mu(\ttheta)
    \quad\text{in }\H_\nu
    \label{eq:measure.synthesis}
\end{align}
is well defined.  For an atomic measure
$\mu=\sum_{j=1}^N c_j\delta_{\ttheta_j}$, it is precisely the finite
network $S_\nu[\mu]=\sum_{j=1}^N c_j\Phi_{\ttheta_j}$.
\end{definition}

\subsection{Approximation by Finite Networks}

The goal is to preserve a nontrivial amount of parameter mass while making
the network output small.  Sampling the polar decomposition of a null
measure gives the required atomic network; normalization rules out the
vacuous construction obtained by shrinking every output coefficient.

\begin{theorem}[Normalized finite-width approximation of a null measure]
\label{thm:finite.null}
Let $\mu_0$ be a nonzero finite complex Radon measure on $\Theta$ satisfying
\begin{align}
    \|\mu_0\|_{\rm TV}=1,
    \qquad
    E_\nu(\mu_0)<\infty,
    \qquad
    S_\nu[\mu_0]=0.
    \label{eq:null.measure.assumptions}
\end{align}
Then, for every $N\ge1$, there is an atomic measure
\begin{align}
    \mu_N=\frac1N\sum_{j=1}^N
    u_j\delta_{\ttheta_j},
    \qquad |u_j|=1,
    \label{eq:finite.null.measure}
\end{align}
such that
\begin{align}
    \|\mu_N\|_{\rm TV}=1,
    \qquad
    \|S_\nu[\mu_N]\|_{\H_\nu}
    \le \frac{E_\nu(\mu_0)^{1/2}}{\sqrt N}.
    \label{eq:finite.null.rate}
\end{align}
Moreover, the atomic measures can be obtained by independent sampling from
$|\mu_0|$ so that
\begin{align}
    \EE\|S_\nu[\mu_N]\|_{\H_\nu}^2
    =\frac{E_\nu(\mu_0)}{N},
    \label{eq:finite.null.expected.rate}
\end{align}
and, almost surely,
\begin{align}
    \mu_N\stackrel{*}{\rightharpoonup}\mu_0,
    \qquad
    S_\nu[\mu_N]\longrightarrow0
    \quad\text{in }\H_\nu.
    \label{eq:finite.null.convergence}
\end{align}
\end{theorem}
\begin{proof}
See \cref{sec:proof.finite.theorem}.  Polar decomposition of $\mu_0$ turns
the synthesis into the mean of a centered $\H_\nu$-valued random variable;
the variance identity gives the $N^{-1/2}$ bound and the Hilbert-space strong
law gives almost-sure convergence.
\end{proof}

The estimate is the Hilbert-space sampling argument underlying the
Maurey--Jones--Barron (MJB) approximation bound
\citep{Pisier1981,Jones1992,Barron1993,kainen.survey},
applied to a nonzero measure whose synthesis is zero.  Its relevance here is
that the continuous structure theorem supplies such null measures.  The
following result makes this connection explicit and shows that
\cref{thm:finite.null} is not conditional on an empty class.

\begin{corollary}[Discretizable ridgelet null elements]
\label{cor:finite.ridgelet.null}
Assume that $\sigma\in\A_{s,t}$ is represented by a continuous function of at most
polynomial growth, and let $\nu$ have compact support.  Then there exist
nonzero $f\in\Sch(\RR^m)$ and $\rho\in\B_{s,t}$ such that
\begin{align}
    \iiprod{\sigma,\rho}=0,
    \qquad
    \gamma_0:=R[f;\rho]\in\ker S\cap\Sch(\Theta).
    \label{eq:schwartz.null.ridgelet}
\end{align}
Consequently,
\begin{align}
    \dd\mu_0(\ttheta)
    :=\frac{\gamma_0(\ttheta)}
    {\|\gamma_0\|_{L^1(\Theta)}}\dd\ttheta
    \label{eq:null.measure.from.ridgelet}
\end{align}
satisfies \eqref{eq:null.measure.assumptions}, and hence admits the
normalized finite-width approximations in \cref{thm:finite.null}.
\end{corollary}
\begin{proof}
See \cref{sec:proof.finite.corollary}.  A compactly supported coefficient vector in
$\ker L_\sigma$ produces a nonzero Schwartz ridgelet null element, whose
normalized density has finite feature energy on a compact data domain.
\end{proof}

\subsection{Exact Finite Null Relations}

Exact finite null relations may also arise from algebraic symmetries of the
activation.  These should be distinguished from the approximation of a
general element of the continuous null space.  Let
$\iota(\aa,b):=(-\aa,-b)$.

\begin{proposition}[Exact finite null relations]
\label{prop:exact.finite.null}
The following statements hold whenever the displayed neurons belong to
$\H_\nu$.
\begin{enumerate}
    \item If $\sigma$ is odd, then
    $\frac12(\delta_{\ttheta}+\delta_{\iota(\ttheta)})\in\ker S_\nu$.
    If $\sigma$ is even, then
    $\frac12(\delta_{\ttheta}-\delta_{\iota(\ttheta)})\in\ker S_\nu$.
    \item If $\sigma(z)=z_+$ and
    $\sum_{j=1}^J c_j\aa_j=0$ and $\sum_{j=1}^Jc_jb_j=0$, then
    \begin{align}
        \sum_{j=1}^Jc_j
        \bigl(\delta_{(\aa_j,b_j)}
        -\delta_{(-\aa_j,-b_j)}\bigr)
        \in\ker S_\nu.
        \label{eq:relu.exact.null}
    \end{align}
\end{enumerate}
\end{proposition}
\begin{proof}
See \cref{sec:proof.exact.finite}.  The first assertion follows from parity,
and the ReLU assertion follows from $z_+-(-z)_+=z$ and the two displayed
affine cancellation conditions.
\end{proof}

The approximation theorem concerns parameters selected to
discretize a fixed continuous null measure.  It does not assert that the
coefficient map associated with every fixed finite collection of distinct
neurons has a nontrivial exact kernel.

\section{Further Developments: Numerical Illustration in Finite Networks}\label{sec:numerical}

The preceding approximation theorem gives an asymptotic finite-width
statement; here we ask whether the same cancellation is visible in a direct
computation.  Numerical integration is itself a discretization: applying a
quadrature rule to the continuous ridgelet parameter distribution replaces
its defining integral by a finite sum of neurons.  We choose a compatible
ridgelet function $\rho$ with $\iiprod{\sigma,\rho}=0$, discretize the
nonzero null element $R[f;\rho]$, and verify that the resulting finite
network represents a function close to zero.  Thus the experiment exhibits
the finite-width trace of the continuous parameter redundancy rather than
an exclusively infinite-width pathology.

We visualize the ridgelet transforms $R[f;\rho](a,b)$ and the corresponding reconstructions $S[R[f;\rho]](x)$ by numerical integration.  See \cref{sec:num} for the experimental details.
For visualization purposes, the data generating function is a 1-dimensional function $f(x) = \sin(2 \pi x)$ for $x \in [-1,1]$, so the ridgelet spectrum $R[f](a,b)$ becomes a 2-dimensional function. We employed $\sigma(t)=\tanh t$ for the activation function and prepared four ridgelet functions $\rho_i$ $(i=1,\ldots,4)$.  They are normalized so that $\iiprod{\sigma,\rho_i}=0$ for $i=1,3$ and $\iiprod{\sigma,\rho_i}=1$ for $i=2,4$.  We additionally use the odd ridgelet function $\rho_{\rm nt}:=\rho_2-\rho_4$, which satisfies $\iiprod{\sigma,\rho_{\rm nt}}=0$ but is not explained by the elementary antipodal null relation for an odd activation.

\begin{figure*}[ht]
    \centering
    \begin{subfigure}[b]{0.2\textwidth}
        \centering
        \includegraphics[width=\textwidth]{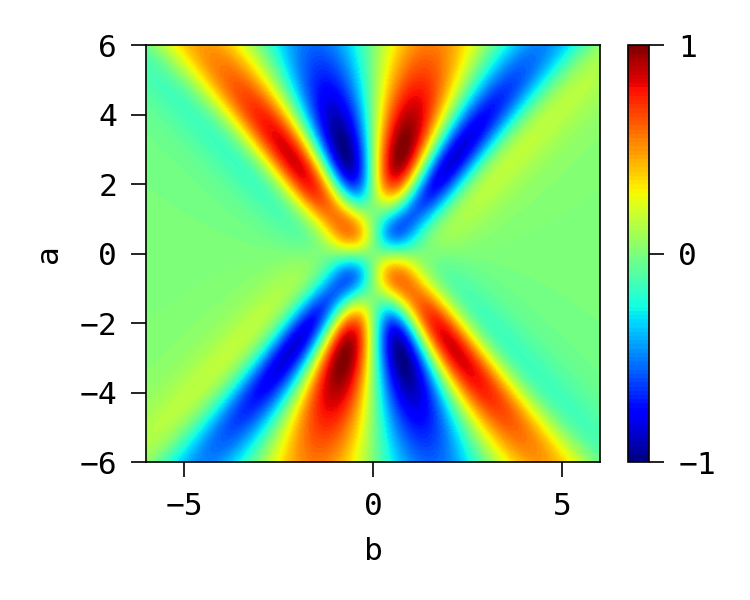}
    \end{subfigure}%
    \begin{subfigure}[b]{0.2\textwidth}
        \centering
        \includegraphics[width=\textwidth]{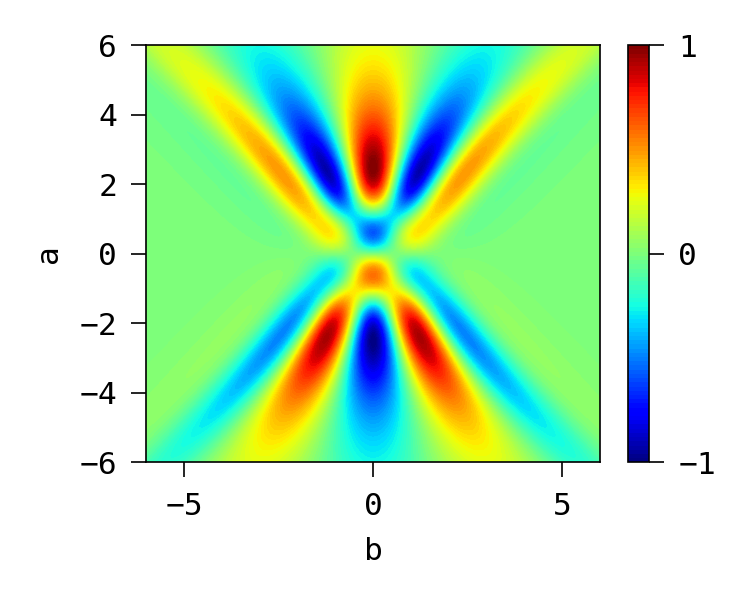}
    \end{subfigure}%
    \begin{subfigure}[b]{0.2\textwidth}
        \centering
        \includegraphics[width=\textwidth]{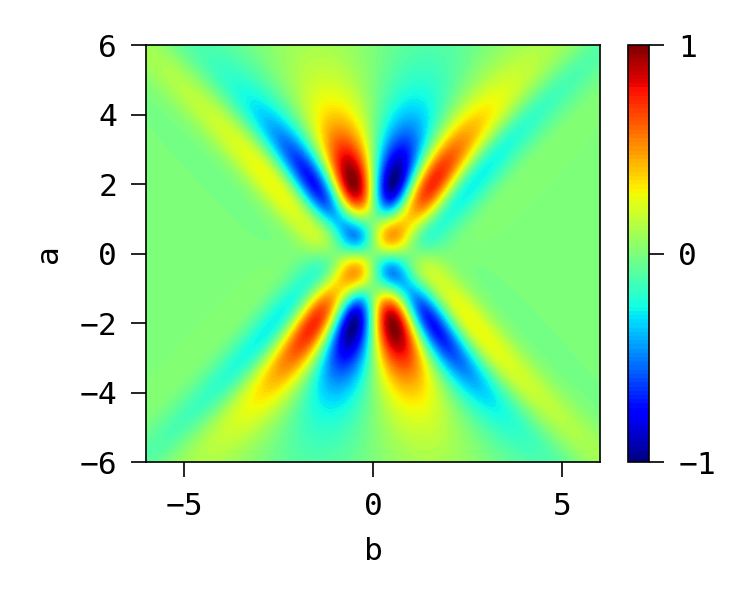}
    \end{subfigure}%
    \begin{subfigure}[b]{0.2\textwidth}
        \centering
        \includegraphics[width=\textwidth]{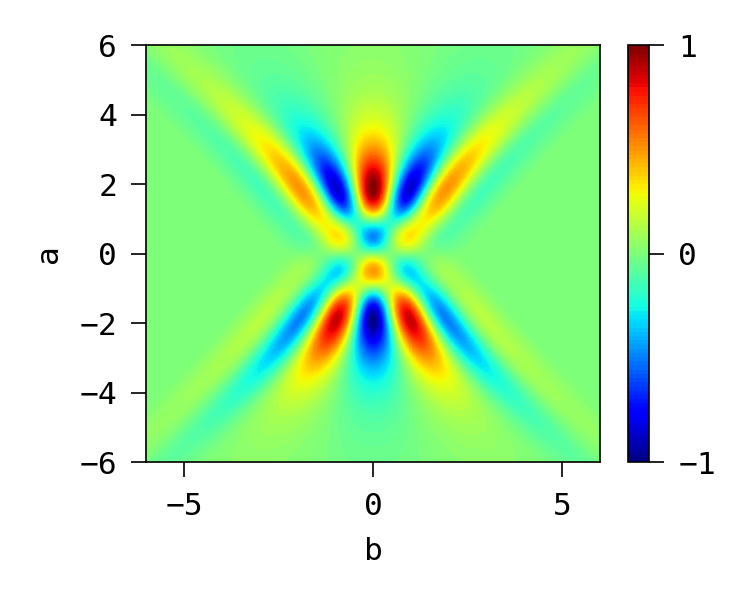}
    \end{subfigure}%
    \begin{subfigure}[b]{0.2\textwidth}
        \centering
        \includegraphics[width=\textwidth]{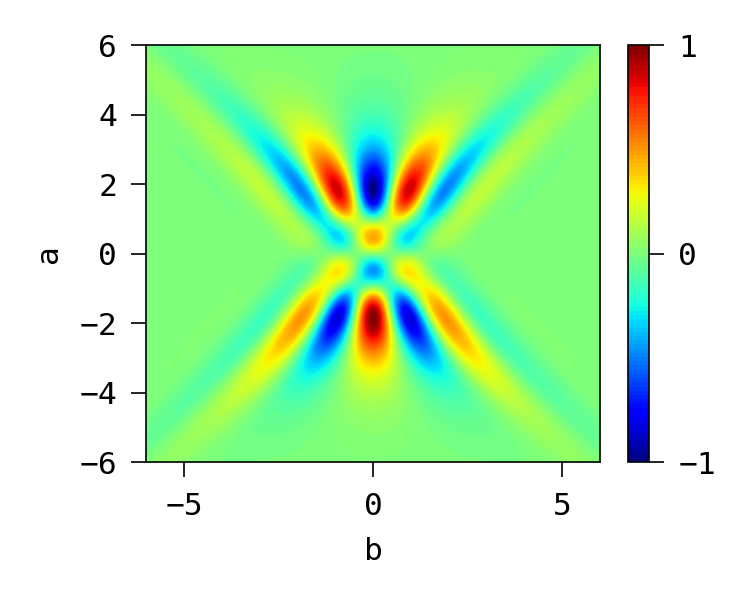}
    \end{subfigure}\\
    \begin{subfigure}[b]{0.2\textwidth}
        \centering
        \includegraphics[width=\textwidth]{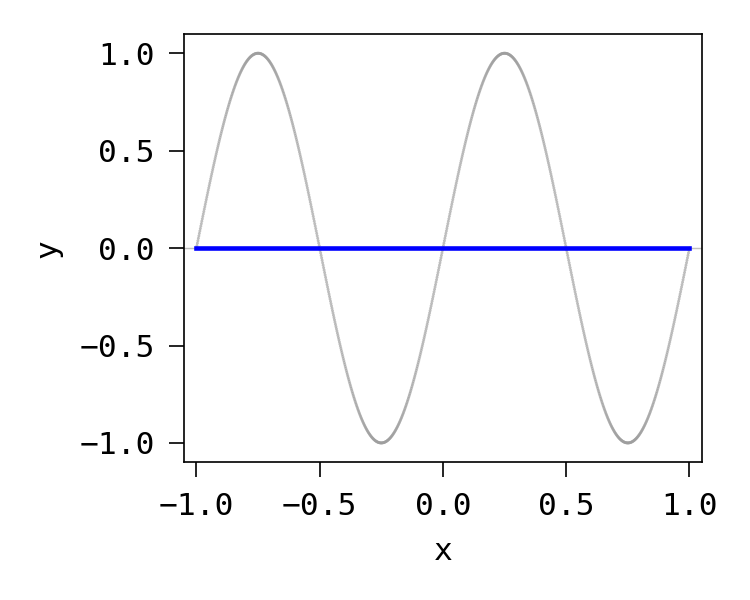}
        \caption{$\iiprod{\sigma,\rho_1}=0$}
        \label{fig:reconst hd1g}
    \end{subfigure}%
    \begin{subfigure}[b]{0.2\textwidth}
        \centering
        \includegraphics[width=\textwidth]{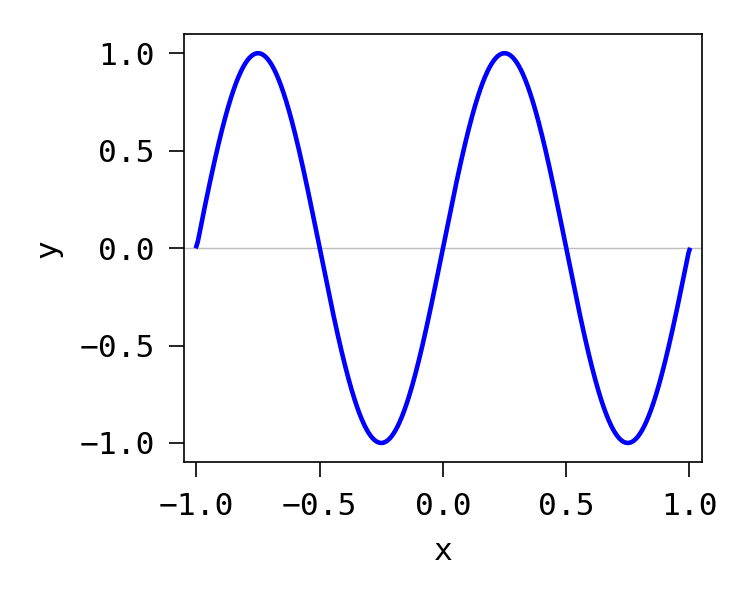}
        \caption{$\iiprod{\sigma,\rho_2}=1$}
        \label{fig:reconst hd2g}
    \end{subfigure}%
    \begin{subfigure}[b]{0.2\textwidth}
        \centering
        \includegraphics[width=\textwidth]{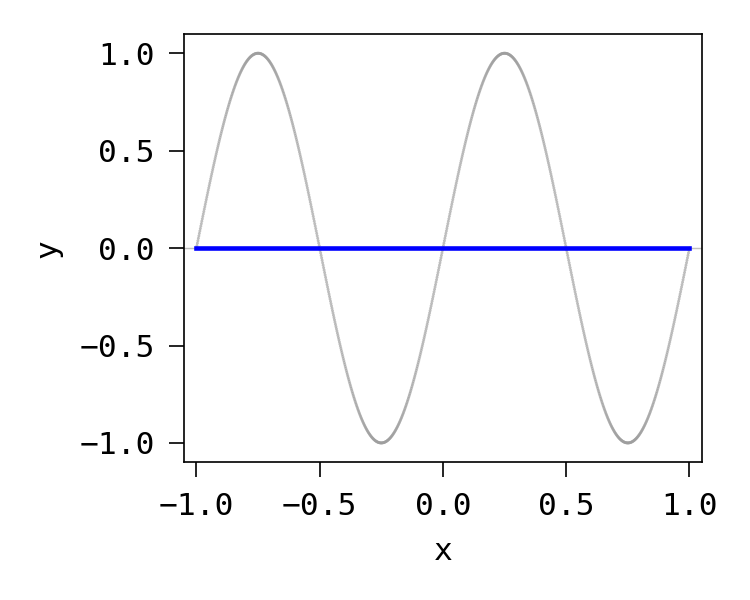}
        \caption{$\iiprod{\sigma,\rho_3}=0$}
        \label{fig:reconst hd3g}
    \end{subfigure}%
    \begin{subfigure}[b]{0.2\textwidth}
        \centering
        \includegraphics[width=\textwidth]{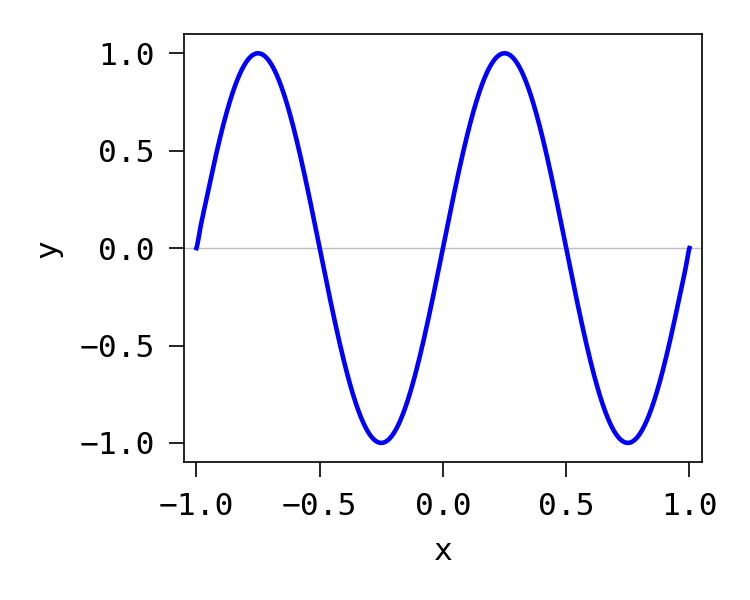}
        \caption{$\iiprod{\sigma,\rho_4}=1$}
        \label{fig:reconst hd4g}
    \end{subfigure}%
    \begin{subfigure}[b]{0.2\textwidth}
        \centering
        \includegraphics[width=\textwidth]{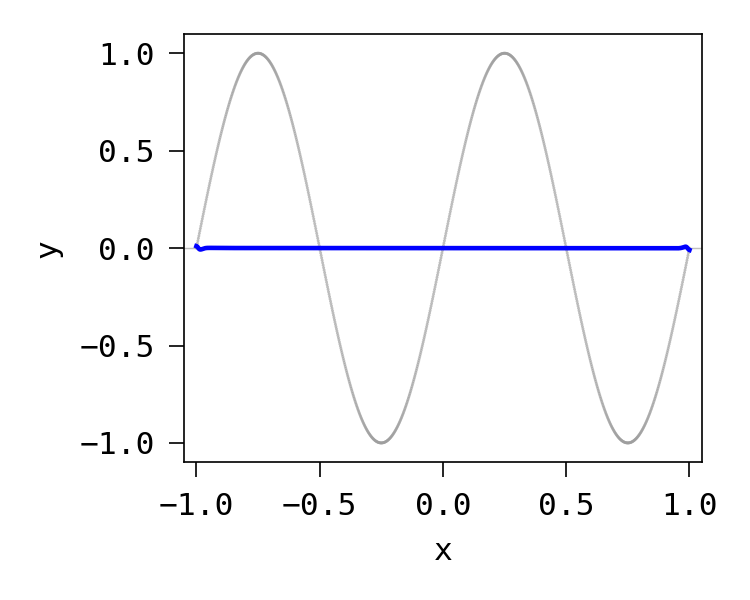}
        \caption{$\iiprod{\sigma,\rho_{\rm nt}}=0$}
        \label{fig:reconst hd2g-hd4g}
    \end{subfigure}\\
    \caption{Ridgelet spectra $R[f;\rho](a,b)$ (up) and reconstruction results $S[R[f;\rho]](x)$ (bottom) for $f(x)=\sin (2 \pi x)\mathbf 1_{[-1,1]}(x)$.  The reconstructions use Gauss--Legendre quadrature, Fourier-normalized coefficients, and finite-box extrapolation; see \cref{sec:num}.}
    \label{fig:null.examples}
\end{figure*}

In \cref{fig:null.examples}, the ridgelet transforms in the upper row are signed and noncompactly supported even though $f$ is compactly supported.  The lower row reproduces $f$ for the admissible choices $\rho_2$ and $\rho_4$, with root-mean-square errors $6.3\times10^{-4}$ and $7.1\times10^{-4}$, respectively.  It vanishes for $\rho_1$ and $\rho_3$ by exact antipodal pairing, and is numerically close to zero for $\rho_{\rm nt}$, with root-mean-square residual $1.1\times10^{-3}$.  Because $\tanh$ is odd while $\rho_1$ and $\rho_3$ are even, the first and third null examples belong to the exact antipodal kernel described in \cref{prop:exact.finite.null}.  By contrast, $\rho_{\rm nt}$ is odd, so its nullity is not a pairwise antipodal cancellation.  Its quadrature therefore provides a genuinely numerical illustration of the nontrivial null relation $R[f;\rho_2]-R[f;\rho_4]$; implementation and error-control details are given in \cref{sec:num}.

The finite-approximation theorem explains the observed cancellation while keeping
the parameter mass nontrivial.  The experiment is illustrative, however;
none of the structural results relies on numerical discretization.

\section{Further Developments: Perturbative Readout of Null-Space Information}\label{sec:readout}

We now turn to a logically separate question: can information in the
nontrivial null component have an observable effect?  If one is allowed to
replace the activation used for decoding, the reconstruction formula makes
readout of a single stored term immediate: a decoder $\sigma'$ satisfying
$\iiprod{\sigma',\rho_i}=1$ reproduces the function paired with $\rho_i$
(and biorthogonal decoders separate several stored functions).  The more
relevant question here is whether the activation can remain fixed while the
parameter distribution is changed.  The idea is simply to execute the
desired output correction in parameter space.  If
$f_0=S[\gamma_{\rm enc}]$, lift $f_i-f_0$ through the canonical right inverse
and add $\delta\gamma_i:=S^\dagger[f_i-f_0]$.
Then $S[\gamma_{\rm enc}+\delta\gamma_i]=f_i$, while the null component is
unchanged because $\delta\gamma_i\in(\ker S)^\perp$.  The tensor description
below makes this mechanism precise for a countable family of encoded
functions and identifies the perturbation as the unique minimum-norm one.

To state the result, for $\tau\in\A_{s,t}$ let $S_\tau$ denote the synthesis
operator with activation $\tau$; thus $S=S_\sigma$.  Over $\CC$, the
reconstruction identity is linear in $f$ and $\tau$ and conjugate-linear in
the ridgelet function $\rho$.

\begin{theorem}[Null-space encoding and parameter perturbations]
\label{thm:perturbative.readout}
There exist sequences $\{\rho_i\}_{i\in\NN}\subset\B_{s,t}$ and
$\{\tau_i\}_{i\in\NN}\subset\A_{s,t}$ such that
\begin{align}
    \{h_{\rho_i}\}_{i\in\NN}
    \text{ is orthonormal in }\H_{s,t},\qquad%
    \iiprod{\sigma,\rho_i}=0,
    \qquad
    \iiprod{\tau_i,\rho_j}=\delta_{ij}.
    \label{eq:encoding.biorthogonality}
\end{align}
Let $f_0\in L^2(\RR^m)$ and let
$(f_i)_{i\in\NN}\in\ell^2(\NN;L^2(\RR^m))$.  Then
\begin{align}
    \gamma_{\rm enc}
    :=S^\dagger[f_0]
      +\sum_{i=1}^\infty R[f_i;\rho_i]
    \quad\text{in }\G_{s,t}
    \label{eq:encoded.parameter}
\end{align}
is well defined and satisfies
\begin{align}
    \left\|\gamma_{\rm enc}-S^\dagger[f_0]\right\|_{\G_{s,t}}^2
    &=\sum_{i=1}^\infty\|f_i\|_{L^2}^2,
    \label{eq:encoding.isometry}\\
    S[\gamma_{\rm enc}]&=f_0,
    \qquad
    S_{\tau_i}[\gamma_{\rm enc}]=f_i.
    \label{eq:activation.readout}
\end{align}
For each $i$, define the parameter perturbation
\begin{align}
    \delta\gamma_i
    &:=S^\dagger
      \big[(S_{\tau_i}-S)[\gamma_{\rm enc}]\big]
      =S^\dagger[f_i-f_0].
    \label{eq:readout.perturbation}
\end{align}
Then the activation remains $\sigma$, while
\begin{align}
    S[\gamma_{\rm enc}+\delta\gamma_i]&=f_i,
    \label{eq:perturbative.readout.output}\\
    (I-P)[\gamma_{\rm enc}+\delta\gamma_i]
    &=(I-P)[\gamma_{\rm enc}].
    \label{eq:perturbative.readout.memory}
\end{align}
Moreover, $\delta\gamma_i$ is the unique minimum-$\G_{s,t}$-norm perturbation
among all $\eta\in\G_{s,t}$ satisfying
$S[\gamma_{\rm enc}+\eta]=f_i$, and
\begin{align}
    \|\delta\gamma_i\|_{\G_{s,t}}
    =\frac{\|f_i-f_0\|_{L^2}}{\sqrt{c_\sigma}}.
    \label{eq:readout.perturbation.norm}
\end{align}
It is also the unique such perturbation that leaves the null component
unchanged as in \eqref{eq:perturbative.readout.memory}.
\end{theorem}
\begin{proof}
See \cref{sec:proof.encoding}.  The construction chooses orthonormal
coefficient vectors in $\ker L_\sigma$ and their Riesz-dual activations; the
$\ell^2$ assumption
is exactly what makes the encoded simple-tensor series converge in the
Bochner space $L^2(\RR^m;\H_{s,t})$.
\end{proof}

Equivalently, the bounded operator
$\Delta_i:=S^\dagger\circ(S_{\tau_i}-S)$ gives
$\delta\gamma_i=\Delta_i[\gamma_{\rm enc}]$.  In the $T$-coordinates
of \eqref{eq:G.Bochner}, this is the rank-one update
\begin{align}
    T[\delta\gamma_i](\xx,\cdot)
    =\bigl(f_i(\xx)-f_0(\xx)\bigr)\frac{h_\sigma}{c_\sigma}.
    \label{eq:readout.coordinate.perturbation}
\end{align}
Thus $\tau_i$ specifies which hidden coordinate is read, while the
perturbed parameter distribution is still synthesized with the original
activation $\sigma$.  The perturbation copies the selected hidden function
into $(\ker S)^\perp$ without deleting any of the information stored in
$\ker S$.  Along the path
$\gamma_i(t):=\gamma_{\rm enc}+t\delta\gamma_i$, $0\le t\le1$, one has
\begin{align}
    S[\gamma_i(t)]=(1-t)f_0+tf_i,
    \qquad
    (I-P)[\gamma_i(t)]=(I-P)[\gamma_{\rm enc}].
    \label{eq:continuous.perturbative.readout}
\end{align}

The square-summability assumption expresses stable Hilbert-space storage.
An arbitrary sequence $(f_i)_{i\in\NN}\subset L^2(\RR^m)$ can still be
encoded by choosing numbers $\alpha_i>0$ such that
$\sum_i\alpha_i^2\|f_i\|_{L^2}^2<\infty$ and replacing $f_i$ in
\eqref{eq:encoded.parameter} by $\alpha_i f_i$.  The $i$th function is then
read by replacing $S_{\tau_i}$ in \eqref{eq:readout.perturbation} with
$S_{\alpha_i^{-1}\tau_i}$.  Such weights always exist, for example
$\alpha_i=2^{-i}(1+\|f_i\|_{L^2})^{-1}$.  The factor $\alpha_i^{-1}$
records the corresponding loss of uniform readout stability.

\section{Discussion}\label{sec:discussion}

The preceding sections separate the exact continuous-width solution theory
from concrete adjoint examples and three further developments: finite-width
approximation, numerical illustration, and perturbative readout of
null-space information.  We now state the distinctions needed to interpret
these developments without extending the claims beyond the proved setting.

\subsection{Continuous and Finite Nullity}

The distinction between exact and approximate nullity is essential.  For a
fixed list $\ttheta_1,\ldots,\ttheta_N$, the linear map
$(c_1,\ldots,c_N)\mapsto\sum_jc_j\Phi_{\ttheta_j}$ can be injective.  In
contrast, \cref{thm:finite.null} selects atoms that discretize a specified
nonzero $\mu_0\in\ker S_\nu$ and shows that the output tends to zero while
the coefficient $\ell^1$-mass remains one.  It therefore gives a normalized
approximate null sequence, not an exact kernel for every fixed finite feature
matrix.  \Cref{prop:exact.finite.null} separately records exact relations
caused by parity and by the affine ambiguity of ReLU.

\subsection{Null-Space Information and Parameter Perturbations}

The information-storage interpretation is made precise by
\cref{thm:perturbative.readout}.  The coefficient vectors can encode a
sequence of functions while remaining invisible under $L_\sigma$, and an
appropriate additive perturbation of the parameter distribution reads a
selected function into the output without changing the stored null
component.  The perturbation is structured and depends on the selected
hidden coordinate; an arbitrary perturbation need not perform this readout.
Whether a particular learning algorithm generates such a coupling remains
a separate dynamical question.

\subsection{Scope and Learning-Theoretic Implications}

Only the projected parameter $P[\gamma]$ affects the represented function,
because $S[P[\gamma]]=S[\gamma]$.  This observation is relevant whenever a
complexity measure is expressed in terms of a parameter norm.  It does not
by itself yield a finite-sample generalization bound: such a result must also
specify the discretization, the admissible hypothesis class, and the relation
between its finite-dimensional norm and the $\G_{s,t}$-norm.  The same caution
applies to comparisons with lazy learning and neural tangent kernels.  These
topics are therefore consequences to be investigated, rather than claims of
the present structure theorem.

\section{Conclusion}

The Fourier expression gives a direct route from the equation
$S[\gamma]=f$ to its ridgelet solutions: separation of variables constructs
a particular solution, and the Hilbert-basis expansion accounts for every
null component.  The abstract reconstruction formula isolates the unitary
factorization and orthogonal solution geometry.  Its neural-network
specialization supplies the activation-adapted Hilbert spaces and
boundedness theorem, realizes reconstruction by ridgelet transforms, and
determines the adjoint, the full affine solution set, and its unique
minimum-norm representative.

The subsequent examples identify adjoint ridgelet functions for standard
activations.  The further developments clarify how the continuous theory
extends beyond the exact general solution: finite-measure null elements
yield normalized width-$N$ approximate null networks at rate $N^{-1/2}$,
whereas exact finite relations require additional activation symmetries;
numerical discretization exhibits this cancellation in finite networks; and
a precisely chosen additive parameter perturbation can make null-space
information visible at the output.

\appendix
\section{Proofs and Auxiliary Calculations}\label{sec:proofs}
This appendix supplies the functional-analytic details deferred from the
main development, following the order in which the corresponding results
appear there.

\subsection{Proof of the Abstract Reconstruction Theorem}
\label{sec:proof.abstract.reconstruction}

This subsection proves \cref{prop:abstract.synthesis}.  By definition,
$T[R_h[f]]=f\otimes h$.  Applying the pointwise functional gives
$S[R_h[f]](x)=L[f(x)h]=L[h]f(x)$, which proves
\eqref{eq:abstract.reconstruction}.  For $\gamma\in\G$ and $f\in\F$,
unitarity and the Riesz identity give
\begin{align}
    \iprod{S[\gamma],f}_{\F}
    &=\int_X L[T[\gamma](x)]\overline{f(x)}\dd\mu(x)\notag\\
    &=\iprod{T[\gamma],f\otimes h_L}_{L^2(X;\H)}
      =\iprod{\gamma,R_{h_L}[f]}_{\G}.
\end{align}
Hence $S^*=R_{h_L}$.  Equation
\eqref{eq:abstract.reconstruction} with $h=h_L$ gives $SS^*=c_LI$, and
unitarity gives the norm identity.

\subsection{Proof of the Abstract Solution Geometry}
\label{sec:proof.abstract.geometry}

This subsection proves \cref{thm:abstract.geometry}.  The identity
$SS^*=c_LI$ in \cref{prop:abstract.synthesis} gives
$SS^\dagger=I$ and $P^2=c_L^{-2}S^*(SS^*)S=P$; self-adjointness is
immediate.  Moreover,
\begin{align}
    \iprod{P[\gamma],\gamma}_{\G}
    =c_L^{-1}\|S[\gamma]\|_{\F}^2,
\end{align}
so $\ker P=\ker S$.  Also $\im P\subset\im S^*$, while
$P[S^*[f]]=S^*[f]$ follows from $SS^*=c_LI$; hence
$\im P=\im S^*$.  The range of a self-adjoint projection is the orthogonal
complement of its kernel, proving the remaining projection identity.
By \eqref{eq:abstract.factorization}, $S[\gamma]=0$ precisely when
$L[T[\gamma](x)]=0$ for almost every $x$; unitarity of $T$ proves
\eqref{eq:abstract.kernel}.  Finally, $SS^\dagger=I$ proves
\eqref{eq:abstract.general.solution}, and its two summands are orthogonal by
$\im S^\dagger=(\ker S)^\perp$.  Pythagoras proves
\eqref{eq:abstract.pythagorean} and uniqueness of the minimizer.

\subsection{Proof of the Abstract Basis Expansion}
\label{sec:proof.abstract.expansion}

This subsection proves \cref{thm:abstract.expansion}.  Let
$u=T[\gamma]$.  The net of finite-rank orthogonal projections indexed by
the finite subsets of $I$ converges strongly to the identity on
$L^2(X;\H)\cong\F\widehat\otimes\H$.  Hence
$u=\sum_{i\in I}e_i\otimes h_i[\gamma]$ in the Hilbert-space sense, with
the stated Parseval identity and unique coefficients.  Applying $T^*$
proves \eqref{eq:abstract.parameter.expansion}.  Since a bounded functional
commutes with the Bochner integrals in \eqref{eq:abstract.coefficients},
$\widetilde L[u]=0$ if and only if all scalar Hilbert-basis coefficients
$L[h_i[\gamma]]$ vanish.  This proves
\eqref{eq:abstract.null.coefficients}.  Expanding each $h_i[\gamma]$ in the
basis $\{k_j\}$ of $\ker L$ gives
\eqref{eq:abstract.double.expansion}; Parseval justifies convergence and
the identification of the two Hilbert sums.

\subsection{Proof of the Activation Hilbert Structure}
\label{sec:proof.activation.hilbert}

This subsection proves \cref{prop:activation.hilbert} and the dual estimate
\eqref{eq:activation.to.dual.bound} used in \cref{thm:bdd.S}.

The map in \eqref{eq:activation.isometry} is an isometry by the definition
of the $\A_{s,t}$-norm.  It is also onto.  Indeed, for any $h\in L^2(\RR)$,
define a tempered distribution $\sigma$ by
\begin{align}
    \sigma^\sharp
    :=\iprod{\partial_\omega}^{t}
      [\iprod{\omega}^{-s}h].
    \label{eq:inverse.activation.isometry}
\end{align}
Then $\sigma\in\A_{s,t}$ and its image under
\eqref{eq:activation.isometry} is $h$.  Transporting the $L^2$ inner
product through this isometric isomorphism proves completeness and hence
the Hilbert-space assertion.

For completeness, the same calculation also proves the claimed
activation--fiber-space duality.  For $h\in\Sch(\RR)$, self-adjointness of the
Bessel operator and Cauchy--Schwarz give
\begin{align}
    \left|\int_\RR h(\omega)\sigma^\sharp(\omega)\dd\omega\right|
    &\le
    \left\|\iprod{\omega}^{-s}
    \iprod{\partial_\omega}^{t}[h]\right\|_{L^2}
    \left\|\iprod{\omega}^{s}
    \iprod{\partial_\omega}^{-t}[\sigma^\sharp]\right\|_{L^2}
\le \|h\|_{\H_{s,t}}\|\sigma\|_{\A_{s,t}}.
    \label{eq:proof.activation.fiber.duality}
\end{align}
Density of $\Sch(\RR)$ in $\H_{s,t}$ then proves
that the pairing extends uniquely to $\H_{s,t}$.  Multiplying by the
prefactor $(2\pi)^{m-1}$ in \eqref{eq:ell.sigma} and taking the supremum over
$\|h\|_{\H_{s,t}}=1$ gives \eqref{eq:activation.to.dual.bound}.  The estimate
is uniform in $\sigma$ and therefore also proves the continuity of
$\sigma\mapsto L_\sigma$ as a map into $\H_{s,t}^*$.

\subsection{Proof of the Weighted Dilation Identity}
\label{sec:proof.weighted.dilation}

This subsection proves \cref{lem:weighted.dilation}.  Tonelli's theorem
reduces the claim to a fixed $\omega$.  For every $\omega\ne0$, the linear
map $\xx\mapsto-\omega\xx$ is invertible and the ordinary
change-of-variables theorem gives $\dd\aa=|\omega|^m\dd\xx$.  The
exceptional slice $\{\omega=0\}$ has Lebesgue measure zero, so it contributes
nothing to either iterated integral.  A second application of Tonelli proves
\eqref{eq:weighted.dilation}.

\subsection{Proof of the Coordinate Transform and Parameter Norm}\label{sec:proof.T}

This subsection supplies the proof details for \cref{prop:T.unitary} and
the compatibility of $T$ with its pointwise formula, including
\eqref{eq:T.L2}, \eqref{eq:G.Bochner}, and
\eqref{eq:T.adjoint.inverse}, which are used in
\cref{lem:fourier,thm:struct}.

We first construct the base $L^2$ transform independently of the graph
domain.  Put
\begin{align}
    \mathcal W_m
    :=L^2\!\left(\RR^m\times\RR,
      (2\pi)^{m-1}|\omega|^m\dd\xx\dd\omega\right).
    \label{eq:weighted.coordinate.space}
\end{align}
For $\gamma\in\Sch(\Theta)$, Plancherel's theorem in $b$ and inverse
Plancherel's theorem in $\aa$ give
\begin{align}
    \int_{\RR^m\times\RR}
    |\checks{\gamma}(\yy,\omega)|^2\dd\yy\dd\omega
    &=(2\pi)^{-m}
      \int_{\RR^m\times\RR}|\gamma^\sharp(\aa,\omega)|^2
      \dd\aa\dd\omega\notag\\
    &=(2\pi)^{1-m}
      \|\gamma\|_{L^2(\Theta)}^2.
    \label{eq:proof.T.plancherel}
\end{align}
For $\omega\ne0$, set $\yy=\omega\xx$.  This slice-wise substitution is
precisely \cref{lem:weighted.dilation}; the exceptional set
$\{\omega=0\}$ has product measure zero.  Hence
\begin{align}
    \int_{\RR^m\times\RR}
    |T_{\rm pt}[\gamma](\xx,\omega)|^2|\omega|^m
    \dd\xx\dd\omega
    &=\int_{\RR^m\times\RR}
      |\checks{\gamma}(\yy,\omega)|^2\dd\yy\dd\omega.
\end{align}
Combining the two identities proves \eqref{eq:T.L2}; equivalently,
$T_{\rm pt}:\Sch(\Theta)\to\mathcal W_m$ is an isometry.

This isometry is onto after completion.  Indeed, for
$v\in\mathcal W_m$, define an $L^2$ function, for $\omega\ne0$, by
\begin{align}
    \checks{\gamma}(\yy,\omega):=v(\yy/\omega,\omega).
    \label{eq:inverse.T}
\end{align}
The definition is independent of representatives almost everywhere:
slice-wise dilation preserves null sets away from $\omega=0$, and that
exceptional slice is null.  Moreover,
\begin{align}
    \int_{\RR^m\times\RR}
      |\checks{\gamma}(\yy,\omega)|^2\dd\yy\dd\omega
    =\int_{\RR^m\times\RR}
      |v(\xx,\omega)|^2|\omega|^m\dd\xx\dd\omega<\infty.
\end{align}
Fourier inversion therefore gives a unique
$\gamma\in L^2(\Theta)$ satisfying $T_0[\gamma]=v$.  Thus
$T_{\rm pt}$ extends uniquely to a unitary map
$T_0:L^2(\Theta)\to\mathcal W_m$.

The first term of the $\H_{s,t}$-norm gives a continuous injection
\begin{align}
    L^2(\RR^m;\H_{s,t})\lhook\joinrel\longrightarrow\mathcal W_m.
\end{align}
Consequently, \eqref{eq:G.space} is the inverse image of the Bochner space
under $T_0$.  For $\gamma\in\G_{s,t}$, identity \eqref{eq:T.L2} shows that
\begin{align}
    \|T[\gamma]\|_{L^2(\RR^m;\H_{s,t})}^2
    =\|\gamma\|_{\G_{s,t}}^2.
\end{align}
Conversely, if $u\in L^2(\RR^m;\H_{s,t})$, regard $u$ as an element of
$\mathcal W_m$ and set $\gamma:=T_0^{-1}[u]$.  Then
$\gamma\in\G_{s,t}$ and $T[\gamma]=u$.  This proves the unitarity and
surjectivity in \eqref{eq:G.Bochner}, and completeness of $\G_{s,t}$ follows.

We next verify the compatibility assertion.  If
$\gamma\in\Sch^T_{s,t}(\Theta)$, then the pointwise formula
$T_{\rm pt}[\gamma]$ represents an element of the Bochner space.  Its image
in $\mathcal W_m$ is, by construction of the extension, exactly
$T_0[\gamma]$.  Hence $\gamma\in\G_{s,t}$ and
$T[\gamma]=T_{\rm pt}[\gamma]$ almost everywhere, proving the asserted
compatibility.

The membership hypothesis is essential.  For example, when $(s,t)=(0,0)$
and $\omega\ne0$, the change of variables $\yy=\omega\xx$ gives
\begin{align}
    \int_{\RR^m}|T_{\rm pt}[\gamma](\xx,\omega)|^2\dd\xx
    =|\omega|^{-m}\int_{\RR^m}
      |\checks{\gamma}(\yy,\omega)|^2\dd\yy.
    \label{eq:schwartz.graph.obstruction}
\end{align}
There are Schwartz parameters for which the last integral has a positive
limit at $\omega=0$.  The unweighted $\omega$-integral on the left then
diverges for every $m\ge1$, so such a parameter does not belong to
$\G_{0,0}$.  Pointwise Schwartz regularity in $\omega$ therefore does not
imply Bochner $L^2$ membership in $\xx$.

It remains to justify the inverse integral formula on the named coordinate
class.  By the definition of $\H_{s,t}$, $\Sch(\RR)$ is dense in
$\H_{s,t}$, while $\Sch(\RR^m)$ is dense in $L^2(\RR^m)$.  Density of
algebraic tensors in the Hilbert tensor product
$L^2(\RR^m)\widehat\otimes\H_{s,t}\cong
L^2(\RR^m;\H_{s,t})$ proves that the class $\mathcal E_{s,t}$ defined in
\cref{sec:parameter.space} is dense.

Take $u\in\mathcal E_{s,t}$.  It is a finite sum of products of Schwartz
functions, so the integral in \eqref{eq:T.adjoint.inverse} is absolutely
convergent.  Let $\gamma_u$ denote its value.  Fourier transformation in
$b$ gives
\begin{align}
    \gamma_u^\sharp(\aa,\omega)
    =|\omega|^m\int_{\RR^m}
      u(\xx,\omega)e^{-i\omega\aa\cdot\xx}\dd\xx.
\end{align}
Inverse Fourier transformation in $\aa$ and the substitution
$\yy=\omega\xx$ yield, for $\omega\ne0$,
\begin{align}
    \checks{\gamma_u}(\yy,\omega)=u(\yy/\omega,\omega).
\end{align}
The calculation following \eqref{eq:inverse.T} shows that
$\gamma_u\in L^2(\Theta)$, while
$T[\gamma_u]=u$ almost everywhere.  Thus $\gamma_u=T^{-1}[u]=T^*[u]$,
which proves \eqref{eq:T.adjoint.inverse} on $\mathcal E_{s,t}$.

For a general $u\in L^2(\RR^m;\H_{s,t})$, the operator $T^*[u]$ is the
$\G_{s,t}$-limit of this formula along approximating elements of
$\mathcal E_{s,t}$.  The integral on the right of
\eqref{eq:T.adjoint.inverse} is not asserted to converge pointwise for an
arbitrary Bochner $L^2$ coordinate.

\subsection{Proof of the Boundedness Theorem}\label{sec:proof.bddS}

This subsection proves \cref{thm:bdd.S}.

Let $h\in\Sch(\RR)$ and $\sigma\in\A_{s,t}$.  Self-adjointness of the
Bessel operator in the distributional dual pairing gives
\begin{align}
    \int_\RR h(\omega)\sigma^\sharp(\omega)\dd\omega
    &=\int_\RR
      \iprod{\partial_\omega}^{t}[h](\omega)
      \iprod{\partial_\omega}^{-t}[\sigma^\sharp](\omega)
      \dd\omega\notag\\
    &=\int_\RR
      \left[\iprod{\omega}^{-s}
      \iprod{\partial_\omega}^{t}[h](\omega)\right]
      \left[\iprod{\omega}^{s}
      \iprod{\partial_\omega}^{-t}[\sigma^\sharp](\omega)\right]
      \dd\omega.
    \label{eq:proof.fiber.duality}
\end{align}
Cauchy--Schwarz proves \eqref{eq:fiber.pairing.bound}; density extends the
pairing to every $h\in\H_{s,t}$.

For $\gamma\in\Sch^T_{s,t}(\Theta)$ and a Schwartz activation, Fourier
inversion in the last variable and
the compatibility statement in \cref{sec:parameter.space} give
\begin{align}
    S[\gamma](\xx)
    &=\frac{1}{2\pi}
      \int_{\RR^m\times\RR}
      \gamma^\sharp(\aa,\omega)\sigma^\sharp(\omega)
      e^{i\omega\aa\cdot\xx}\dd\aa\dd\omega\notag\\
    &=(2\pi)^{m-1}\int_\RR
      \checks{\gamma}(\omega\xx,\omega)
      \sigma^\sharp(\omega)\dd\omega,
    \label{eq:proof.S.T}
\end{align}
which is \eqref{eq:S.T}.  The same identity holds for
$\sigma\in\A_{s,t}$ by distributional duality whenever the classical
pairing is defined.  For an arbitrary $\gamma\in\G_{s,t}$, define
$S[\gamma]=\widetilde L_\sigma T[\gamma]$ as in \eqref{eq:S.T}.  Applying
\eqref{eq:fiber.pairing.bound} at almost every $\xx$ then yields
\begin{align}
    \|S[\gamma]\|_{L^2(\RR^m)}^2
    &\le (2\pi)^{2m-2}\|\sigma\|_{\A_{s,t}}^2
      \int_{\RR^m}
      \|T[\gamma](\xx,\cdot)\|_{\H_{s,t}}^2\dd\xx\notag\\
    &=(2\pi)^{2m-2}\|\sigma\|_{\A_{s,t}}^2
      \|\gamma\|_{\G_{s,t}}^2.
\end{align}
Thus \eqref{eq:S.T} defines a bounded operator on $\G_{s,t}$, and
\eqref{eq:S.bound} follows.  The calculation above shows that it agrees with
the classical integral representation whenever both sides are classically
defined.

\subsection{Proof of the Fourier Formulas and Reconstruction Theorem}\label{sec:proof.fourier}

This subsection proves \cref{lem:fourier,thm:reconst}.

We first take $f$ and $\rho$ to be Schwartz functions.  Fourier transform
in $b$ gives
\begin{align}
    R[f;\rho]^\sharp(\aa,\omega)
    &=\int_{\RR^m\times\RR}
      f(\xx)\overline{\rho(\aa\cdot\xx-b)}e^{-ib\omega}
      \dd b\dd\xx\notag\\
    &=\widehat f(\omega\aa)
      \overline{\rho^\sharp(\omega)},
    \label{eq:proof.fourierR}
\end{align}
which proves \eqref{eq:fourierR} on the dense subspace.  Inverting the
$\aa$-Fourier transform at $\omega\xx$ and setting
$\xxi=\omega\aa$ yield
\begin{align}
    T[R[f;\rho]](\xx,\omega)
    &=\frac{1}{(2\pi)^m}
      \int_{\RR^m}\widehat f(\omega\aa)
      \overline{\rho^\sharp(\omega)}e^{i\omega\aa\cdot\xx}
      \dd\aa\notag\\
    &=f(\xx)|\omega|^{-m}
      \overline{\rho^\sharp(\omega)}
      =f(\xx)h_\rho(\omega).
    \label{eq:proof.R.T}
\end{align}
It follows from the unitary identification \eqref{eq:G.Bochner} that
\begin{align}
    \|R[f;\rho]\|_{\G_{s,t}}^2
    =\|f\|_{L^2(\RR^m)}^2\|h_\rho\|_{\H_{s,t}}^2.
    \label{eq:proof.R.Gnorm}
\end{align}
In particular, the $L^2$ part of this identity is
\begin{align}
    \|R[f;\rho]\|_{L^2(\RR^m\times\RR)}^2
    =\|f\|_{L^2(\RR^m)}^2\|\rho\|_{L_m^2(\RR)}^2.
    \label{eq:R.L2.isometry}
\end{align}
These identities extend \eqref{eq:fourierR} and \eqref{eq:R.T} to
$f\in L^2$ and $\rho\in\B_{s,t}$.

For the Fourier expression of $S$, take the $\xx$-Fourier transform of
\eqref{eq:S.T}.  The change of variables $\yy=\omega\xx$ gives
\begin{align}
    \widehat{T[\gamma]}(\xxi,\omega)
    &=\int_{\RR^m}\checks{\gamma}(\omega\xx,\omega)
      e^{-i\xx\cdot\xxi}\dd\xx\notag\\
    &=|\omega|^{-m}\gamma^\sharp(\xxi/\omega,\omega).
    \label{eq:proof.T.fourier}
\end{align}
Substitution into \eqref{eq:S.T} proves \eqref{eq:fourierS}.  The identity
extends from the dense subspace by \cref{thm:bdd.S}.

Finally, \eqref{eq:S.T} and \eqref{eq:proof.R.T} give
\begin{align}
    S[R[f;\rho]](\xx)
    &=(2\pi)^{m-1}f(\xx)
      \int_\RR h_\rho(\omega)\sigma^\sharp(\omega)\dd\omega\notag\\
    &=\iiprod{\sigma,\rho}f(\xx),
\end{align}
proving \cref{thm:reconst}.

\subsection{Proof of the Adjoint Formula}\label{sec:proof.adjS}

This subsection proves \cref{lem:adj.S}.

By \eqref{eq:fiber.pairing.bound}, $L_\sigma$ is a bounded linear
functional on $\H_{s,t}$, so the Riesz representer $h_\sigma$ in
\eqref{eq:h.sigma} exists and is unique.  The definition
\eqref{eq:sigma.star} gives $h_{\sigma_*}=h_\sigma$ and hence, by
\eqref{eq:R.T},
\begin{align}
    T[R[f;\sigma_*]](\xx,\omega)
    =f(\xx)h_\sigma(\omega).
    \label{eq:proof.adjoint.T}
\end{align}
For $f\in L^2(\RR^m)$ and $\gamma\in\G_{s,t}$, the convention that inner
products are linear in the first argument yields
\begin{align}
    \iprod{R[f;\sigma_*],\gamma}_{\G_{s,t}}
    &=\int_{\RR^m}
      f(\xx)\iprod{h_\sigma,
      T[\gamma](\xx,\cdot)}_{\H_{s,t}}\dd\xx\notag\\
    &=\int_{\RR^m}
      f(\xx)\overline{L_\sigma[
      T[\gamma](\xx,\cdot)]}\dd\xx\notag\\
    &=\iprod{f,S[\gamma]}_{L^2(\RR^m)}.
    \label{eq:proof.adjoint.identity}
\end{align}
Thus $S^*[f]=R[f;\sigma_*]$.

Applying $S$ to \eqref{eq:proof.adjoint.T}, we obtain
\begin{align}
    S[S^*[f]]
    =fL_\sigma[h_\sigma]
    =f\iprod{h_\sigma,h_\sigma}_{\H_{s,t}}
    =c_\sigma f.
\end{align}
Similarly,
\begin{align}
    \|S^*[f]\|_{\G_{s,t}}^2
    =\|f\|_{L^2}^2\|h_\sigma\|_{\H_{s,t}}^2
    =c_\sigma\|f\|_{L^2}^2.
\end{align}
Since $h_{\sigma_*}=h_\sigma$, the definition of the Fourier pairing gives
\begin{align}
    \iiprod{\sigma,\sigma_*}
    =L_\sigma[h_\sigma]=c_\sigma.
\end{align}
This proves \eqref{eq:adjoint.identities}.

\subsection{Proof of the Projection Identities}

This subsection proves \eqref{eq:projection.properties}, which is used in
\cref{thm:struct,thm:perturbative.readout}.

For completeness, $P=c_\sigma^{-1}S^*S$ is self-adjoint and
\begin{align}
    P^2
    =c_\sigma^{-2}S^*(SS^*)S
    =c_\sigma^{-1}S^*S=P.
\end{align}
Moreover, $P[\gamma]=0$ if and only if
\begin{align}
    0=\iprod{P[\gamma],\gamma}_{\G_{s,t}}
    =c_\sigma^{-1}\|S[\gamma]\|_{L^2}^2,
\end{align}
so $\ker P=\ker S$.  The remaining statements in
\eqref{eq:projection.properties} follow from the standard orthogonal
decomposition associated with a self-adjoint projection.

\subsection{Proof of the Structure Theorem}\label{sec:proof.struct}

This subsection proves \cref{thm:struct}.

Equation \eqref{eq:S.T} can be written as
\begin{align}
    S[\gamma](\xx)
    =L_\sigma[T[\gamma](\xx,\cdot)].
    \label{eq:proof.S.fiber}
\end{align}
This proves the equivalence in part~1.  Since $T$ is unitary onto
$L^2(\RR^m;\H_{s,t})$, the same equivalence gives
\eqref{eq:null.tensor.characterization}.

Let $\gamma_0\in\ker S$ and set
$u_0:=T[\gamma_0]\in L^2(\RR^m;\H_{s,t})$.  The Hilbert-space-valued
orthogonal expansion with respect to the orthonormal basis
$\{e_i\}_{i\in\NN}$ gives
\begin{align}
    u_0(\xx,\omega)
    =\sum_{i=1}^\infty e_i(\xx)h_i(\omega)
    \quad\text{in }L^2(\RR^m;\H_{s,t}),
    \label{eq:proof.Bochner.series}
\end{align}
where $h_i$ is given by \eqref{eq:hi.coefficient}.  Parseval's identity is
\begin{align}
    \sum_{i=1}^\infty\|h_i\|_{\H_{s,t}}^2
    =\|u_0\|_{L^2(\RR^m;\H_{s,t})}^2
    =\|\gamma_0\|_{\G_{s,t}}^2.
    \label{eq:proof.Bochner.parseval}
\end{align}
Define $\rho_i$ by \eqref{eq:rhoi.coefficient}.  Then
$h_{\rho_i}=h_i$, so \eqref{eq:R.T} gives
\begin{align}
    T[R[e_i;\rho_i]](\xx,\omega)
    =e_i(\xx)h_i(\omega).
\end{align}
The unitary identification \eqref{eq:G.Bochner} turns
\eqref{eq:proof.Bochner.series} into the convergent series
\eqref{eq:null.ridgelet.series} in $\G_{s,t}$.

Since $S$ is bounded, applying it to the partial sums is legitimate:
\begin{align}
    0=S[\gamma_0]
    =\sum_{i=1}^\infty
      \iiprod{\sigma,\rho_i}e_i
    \quad\text{in }L^2(\RR^m).
\end{align}
Completeness and orthonormality of $\{e_i\}$ imply
$\iiprod{\sigma,\rho_i}=0$ for every $i$.  Conversely, any series of the
form \eqref{eq:null.ridgelet.series} that converges in $\G_{s,t}$ and satisfies
these orthogonality conditions is mapped to zero.  Uniqueness follows from
the uniqueness of the Bochner coefficients $h_i$; formula
\eqref{eq:rhoi.coefficient} then fixes each $\rho_i$ without a separate
normalization or phase choice.

Finally, \eqref{eq:abstract.general.solution} gives the canonical affine
description
\begin{align}
    \{\gamma\in\G_{s,t}:S[\gamma]=f\}
    =S^\dagger[f]+\ker S.
\end{align}

If $\rho_0\in\B_{s,t}$ satisfies
$\iiprod{\sigma,\rho_0}=1$, then \cref{thm:reconst} gives
$S[R[f;\rho_0]]=f$.  Hence
\begin{align}
    R[f;\rho_0]-S^\dagger[f]\in\ker S,
\end{align}
which proves the equality of the two affine descriptions in
\eqref{eq:general.solution}.  Applying the unique null expansion
\eqref{eq:null.ridgelet.series} to
$\gamma-R[f;\rho_0]$ gives \eqref{eq:general.solution.series}, including
its convergence and uniqueness.  This explicitly connects the structure
theorem to the Fourier separation-of-variables solution without repeating
the Pythagorean argument of \cref{thm:abstract.geometry}.

\subsection{Proof of the Null-Space Encoding and Parameter-Perturbation Theorem}
\label{sec:proof.encoding}

This subsection proves \cref{thm:perturbative.readout}.

Let
\begin{align}
    \mathcal D_0:=C_c^\infty(\RR\setminus\{0\}).
    \label{eq:D0.definition}
\end{align}
This is an infinite-dimensional subspace of $\H_{s,t}$.  The kernel of the
restriction of $L_\sigma$ to $\mathcal D_0$ has codimension at most one and
is therefore infinite-dimensional.  Choose an $\H_{s,t}$-orthonormal sequence
\begin{align}
    \{h_i\}_{i\in\NN}
    \subset \mathcal D_0\cap\ker L_\sigma.
    \label{eq:encoding.h.sequence}
\end{align}
Define
\begin{align}
    \rho_i^\sharp(\omega)
    :=|\omega|^m\overline{h_i(\omega)}.
    \label{eq:encoding.rho.construction}
\end{align}
Then $h_{\rho_i}=h_i$, so $\rho_i\in\B_{s,t}$ and
$\iiprod{\sigma,\rho_i}=L_\sigma[h_i]=0$.

We next construct a dual activation for each coordinate.  Set
\begin{align}
    \tau_i^\sharp(\omega)
    &:=
    |\omega|^m\overline{h_i(\omega)}+(2\pi)^{1-m}
    \iprod{\partial_\omega}^{t}
    \left[
      \iprod{\omega}^{-2s}
      \overline{\iprod{\partial_\omega}^{t}[h_i](\omega)}
    \right].
    \label{eq:dual.activation.construction}
\end{align}
Because $h_i$ is smooth, compactly supported, and supported away from the
origin, the first term is a test function.  The Bessel operator preserves
the Schwartz space, so the second term is Schwartz as well.  Hence
$\tau_i^\sharp\in\Sch(\RR)$ and $\tau_i\in\Sch(\RR)\subset\A_{s,t}$.
Self-adjointness of the Bessel operator gives, first for $h\in\Sch(\RR)$
and then by density for every $h\in\H_{s,t}$,
\begin{align}
    L_{\tau_i}[h]
    &=(2\pi)^{m-1}\int_\RR
      h(\omega)\overline{h_i(\omega)}|\omega|^m\dd\omega+
      \int_\RR
      \iprod{\partial_\omega}^{t}[h](\omega)
      \overline{\iprod{\partial_\omega}^{t}[h_i](\omega)}
      \iprod{\omega}^{-2s}\dd\omega=\iprod{h,h_i}_{\H_{s,t}}.
    \label{eq:dual.activation.riesz}
\end{align}
Thus the Riesz representer of $L_{\tau_i}$ is $h_i$, and
\begin{align}
    \iiprod{\tau_i,\rho_j}
    =L_{\tau_i}[h_j]
    =\iprod{h_j,h_i}_{\H_{s,t}}
    =\delta_{ij},
\end{align}
which proves \eqref{eq:encoding.biorthogonality}.

Since $\{h_i\}$ is orthonormal and $(f_i)\in\ell^2$, the series
\begin{align}
    \sum_{i=1}^\infty f_i(\xx)h_i(\omega)
    \label{eq:encoded.coordinate.series}
\end{align}
converges in $L^2(\RR^m;\H_{s,t})$ and its squared norm is
$\sum_i\|f_i\|_{L^2}^2$.  Equations \eqref{eq:G.Bochner} and
\eqref{eq:R.T} therefore prove the convergence in
\eqref{eq:encoded.parameter} and the norm identity
\eqref{eq:encoding.isometry}.  Moreover,
$h_i\perp h_\sigma$ because $h_i\in\ker L_\sigma$.  Consequently,
\begin{align}
    S\left[\sum_{j=1}^\infty R[f_j;\rho_j]\right]&=0,\\
    S_{\tau_i}[S^\dagger[f_0]]&=0,\\
    S_{\tau_i}\left[\sum_{j=1}^\infty R[f_j;\rho_j]\right]
    &=f_i.
\end{align}
Here the series may be passed through the synthesis operators because they
are bounded.  Together with $SS^\dagger=I$, these identities prove
\eqref{eq:activation.readout}.

By definition and \eqref{eq:activation.readout},
\begin{align}
    \delta\gamma_i=S^\dagger[f_i-f_0].
\end{align}
It belongs to $\im S^\dagger=(\ker S)^\perp$.  Hence
\begin{align}
    S[\gamma_{\rm enc}+\delta\gamma_i]
    &=f_0+S[S^\dagger[f_i-f_0]]=f_i,\\
    (I-P)[\gamma_{\rm enc}+\delta\gamma_i]
    &=(I-P)[\gamma_{\rm enc}],
\end{align}
proving \eqref{eq:perturbative.readout.output} and
\eqref{eq:perturbative.readout.memory}.

Finally, every perturbation $\eta$ with
$S[\gamma_{\rm enc}+\eta]=f_i$ satisfies
\begin{align}
    \eta=S^\dagger[f_i-f_0]+\eta_0,
    \qquad \eta_0\in\ker S,
\end{align}
by \cref{thm:struct}.  Orthogonality and \eqref{eq:adjoint.identities} give
\begin{align}
    \|\eta\|_{\G_{s,t}}^2
    =\frac{\|f_i-f_0\|_{L^2}^2}{c_\sigma}
     +\|\eta_0\|_{\G_{s,t}}^2.
    \label{eq:proof.minimum.perturbation}
\end{align}
This proves the minimum-norm claim and
\eqref{eq:readout.perturbation.norm}.  If the null component must remain
fixed, then $(I-P)[\eta]=0$, forcing $\eta_0=0$ and proving the final
uniqueness assertion.

\subsection{Derivation of the Adjoint Examples}
\label{sec:proof.adjoint.examples}

This subsection derives \eqref{eq:adjoint.ridgelet.examples} and
\eqref{eq:gaussian.adjoint.closed} from the adjoint construction in
\cref{lem:adj.S}.

Let $r_\sigma=\overline{h_\sigma}$.  Moving the Bessel operator in the
second term of \eqref{eq:h.sigma.weak} by self-adjointness gives
\begin{align}
    &C_m\int_\RR h(\omega)|\omega|^m
      r_\sigma(\omega)\dd\omega+
      \int_\RR h(\omega)
      \iprod{\partial_\omega}^{t}
      \left[
        \iprod{\omega}^{-2s}
        \iprod{\partial_\omega}^{t}[r_\sigma](\omega)
      \right]\dd\omega
      =C_m\int_\RR h(\omega)\sigma^\sharp(\omega)\dd\omega.
    \label{eq:proof.adjoint.riesz.operator}
\end{align}
Since this holds for every test function $h$, it is exactly
\eqref{eq:adjoint.riesz.operator}.  When $(s,t)=(0,2)$,
$\iprod{\partial_\omega}^{2}\iprod{\partial_\omega}^{2}
=(1-\partial_\omega^2)^2$, which gives \eqref{eq:common.adjoint.operator}
and \eqref{eq:adjoint.ridgelet.examples}.  Membership in $\A_{0,2}$ follows
directly from
$\sigma_\bullet/\iprod{\cdot}^{2}\in L^2(\RR)$ for each activation in
\eqref{eq:example.activations}.

For completeness, we verify the Fourier transforms in
\eqref{eq:activation.fourier.examples}.  In the convention
$\phi^\sharp(\omega)=\int\phi(z)e^{-iz\omega}\dd z$, the generalized
Fourier transforms of the Heaviside function and its first integral give
\begin{align}
    \mathbf1_{[0,\infty)}^\sharp
    &=\pi\delta_0-i\,\pv\frac1\omega,\notag\\
    (z_+)^\sharp
    &=-\fp\frac1{\omega^2}+i\pi\delta_0'.
    \label{eq:proof.relu.fourier}
\end{align}
These formulas are the sign-adjusted versions of the formulas tabulated by
\citet[pp.~359--360, Entries~22--23]{GelfandShilov}, whose convention uses
$e^{+iz\omega}$.

The standard Gaussian integral gives
\begin{align}
    \left((2\pi)^{-1/2}e^{-z^2/2}\right)^\sharp
    =e^{-\omega^2/2}.
\end{align}
Since $\Phi_{\rm N}'=\sigma_{\rm G}$ and
$\Phi_{\rm N}=\frac12+\frac12\erf(z/\sqrt2)$, distributional
differentiation and the constant term give
\begin{align}
    \Phi_{\rm N}^\sharp
    =\pi\delta_0-i e^{-\omega^2/2}\pv\frac1\omega.
\end{align}
Finally,
\begin{align}
    (\tanh z)'&=\frac1{\cosh^2 z},
    &
    \left(\frac1{\cosh^2 z}\right)^\sharp
    &=\frac{\pi\omega}{\sinh(\pi\omega/2)}.
\end{align}
Dividing the latter identity by $i\omega$ in the odd tempered-distribution
class yields
\begin{align}
    (\tanh z)^\sharp
    =-i\pi\,\pv\frac1{\sinh(\pi\omega/2)}.
\end{align}

For $(s,t)=(0,0)$, \eqref{eq:adjoint.riesz.operator} reduces to
\begin{align}
    (1+C_m|\omega|^m)r_{\sigma_{\rm G}}(\omega)
    =C_m e^{-\omega^2/2}.
\end{align}
Multiplication by $|\omega|^m$ proves
\eqref{eq:gaussian.adjoint.closed}.

\section{Proofs for Finite-Width Approximation}\label{sec:proof.finite}
This appendix proves the approximation and exact-relation results of
\cref{sec:finite}, and then records a complementary uniform quadrature
estimate on truncated parameter domains.

\subsection{Proof of the Finite-Width Approximation Theorem}
\label{sec:proof.finite.theorem}

This subsection proves \cref{thm:finite.null}.

Let
\begin{align}
    \dd\mu_0(\ttheta)=u(\ttheta)\dd|\mu_0|(\ttheta),
    \qquad |u(\ttheta)|=1
    \quad\text{for }|\mu_0|\text{-almost every }\ttheta,
    \label{eq:polar.null.measure}
\end{align}
be the polar decomposition.  Since $\|\mu_0\|_{\rm TV}=1$, the measure
$|\mu_0|$ is a probability measure.  Let
$\ttheta_1,\ldots,\ttheta_N$ be independent samples from $|\mu_0|$ and
define $\mu_N$ by \eqref{eq:finite.null.measure} with
$u_j=u(\ttheta_j)$.  Repeated samples at an atom carry the same phase, so
combining them does not reduce the total variation.  Hence
$\|\mu_N\|_{\rm TV}=1$ almost surely.

Put $Y_j:=u(\ttheta_j)\Phi_{\ttheta_j}\in\H_\nu$.  The feature-energy
assumption gives $\EE\|Y_j\|_{\H_\nu}^2=E_\nu(\mu_0)<\infty$, while
\begin{align}
    \EE Y_j
    =\int_\Theta\Phi_\ttheta\dd\mu_0(\ttheta)
    =S_\nu[\mu_0]=0.
\end{align}
Independence therefore yields
\begin{align}
    \EE\|S_\nu[\mu_N]\|_{\H_\nu}^2
    &=\EE\left\|\frac1N\sum_{j=1}^NY_j\right\|_{\H_\nu}^2=\frac1{N^2}\sum_{j=1}^N
      \EE\|Y_j\|_{\H_\nu}^2
      =\frac{E_\nu(\mu_0)}{N}.
    \label{eq:proof.finite.expected}
\end{align}
Thus at least one realization satisfies \eqref{eq:finite.null.rate}.

The strong law of large numbers for separable Hilbert spaces gives
$N^{-1}\sum_{j=1}^NY_j\to0$ in $\H_\nu$ almost surely.  Moreover, for
each $g\in C_0(\Theta)$, the scalar strong law gives
\begin{align}
    \int_\Theta g\dd\mu_N
    =\frac1N\sum_{j=1}^N u(\ttheta_j)g(\ttheta_j)
    \longrightarrow
    \int_\Theta g\dd\mu_0.
\end{align}
Because $C_0(\Theta)$ is separable, the convergence holds simultaneously
on a countable dense subset and hence on all of $C_0(\Theta)$, using the
uniform total-variation bound.  This proves the weak-* convergence in
\eqref{eq:finite.null.convergence}.

\subsection{Proof of the Discretizable Ridgelet Corollary}
\label{sec:proof.finite.corollary}

This subsection proves \cref{cor:finite.ridgelet.null}.

The test-function space $\mathcal D_0$ introduced in
\eqref{eq:D0.definition} is an infinite-dimensional subspace of $\H_{s,t}$.
The restriction of the
continuous functional $L_\sigma$ to $\mathcal D_0$ has an
infinite-dimensional kernel.  Choose a nonzero
$h\in\mathcal D_0\cap\ker L_\sigma$ and define
\begin{align}
    \rho^\sharp(\omega)
    :=|\omega|^m\overline{h(\omega)}.
    \label{eq:compact.rho.construction}
\end{align}
Then $h_\rho=h$, so $\rho\in\B_{s,t}$ and
\begin{align}
    \iiprod{\sigma,\rho}=L_\sigma[h]=0.
\end{align}

Choose a nonzero $f\in\Sch(\RR^m)$ with
$\widehat f\in C_c^\infty(\RR^m)$.  The Fourier formula for the ridgelet
transform gives
\begin{align}
    \gamma_0^\sharp(\aa,\omega)
    =\widehat f(\omega\aa)
      \overline{\rho^\sharp(\omega)}.
    \label{eq:compact.null.fourier}
\end{align}
The $\omega$-support of $\rho^\sharp$ is compact and separated from zero.
Together with the compact support of $\widehat f$, this shows that the
right-hand side is a nonzero element of
$C_c^\infty(\RR^m\times\RR)$.  Its inverse Fourier transform in $\omega$
is compactly supported in $\aa$ and rapidly decreasing in $b$, uniformly
with all derivatives.  Thus $\gamma_0\in\Sch(\Theta)$ and
$\gamma_0\ne0$.  The reconstruction formula gives
\begin{align}
    S[\gamma_0]
    =S[R[f;\rho]]
    =\iiprod{\sigma,\rho}f=0.
\end{align}
Because $\gamma_0$ is Schwartz and $\sigma$ is continuous with polynomial
growth, its classical synthesis integral is locally uniformly convergent and
defines a continuous function of $\xx$.  It agrees with the $L^2$ synthesis
above and is therefore identically zero, not merely zero almost everywhere.
Consequently, its restriction represents the zero element of $\H_\nu$ even
when $\nu$ is singular with respect to Lebesgue measure.

It remains to check the feature energy.  If $K:=\supp\nu$ is compact and
$|\sigma(t)|\le C(1+|t|)^r$, then
\begin{align}
    \|\Phi_{(\aa,b)}\|_{\H_\nu}^2
    \le C_K(1+|\aa|+|b|)^{2r}.
    \label{eq:polynomial.feature.bound}
\end{align}
All polynomial moments of $|\gamma_0|$ are finite because $\gamma_0$ is
Schwartz.  Hence the measure in \eqref{eq:null.measure.from.ridgelet} has
finite feature energy.  Its total variation is one by construction, and
its synthesis is zero, completing the proof.

\subsection{Proof of the Exact Finite Null Relations}
\label{sec:proof.exact.finite}

This subsection proves \cref{prop:exact.finite.null}.

For $z=\aa\cdot\xx-b$, the antipodal neuron is
$\Phi_{\iota(\aa,b)}(\xx)=\sigma(-z)$.  The first assertion follows
immediately when $\sigma(-z)=-\sigma(z)$ or
$\sigma(-z)=\sigma(z)$, respectively.  For ReLU,
\begin{align}
    \sigma(z)-\sigma(-z)=z.
\end{align}
Consequently, the synthesis of the measure in
\eqref{eq:relu.exact.null} is
\begin{align}
    \sum_{j=1}^Jc_j(\aa_j\cdot\xx-b_j)
    =\left(\sum_{j=1}^Jc_j\aa_j\right)\cdot\xx
     -\sum_{j=1}^Jc_jb_j=0.
\end{align}

\subsection{Uniform Monte Carlo Quadrature on a Truncated Parameter Domain}

The numerical illustration uses uniform sampling on a bounded box rather
than the total-variation sampling in \cref{thm:finite.null}.  The following
estimate separates truncation and sampling errors.

\begin{lemma}\label{lem:truncated.mc}
Let $B\subset\Theta$ have finite positive Lebesgue measure $|B|$, and let
$\gamma:B\to\CC$ satisfy
\begin{align}
    \int_B |\gamma(\ttheta)|^2
    \|\Phi_\ttheta\|_{\H_\nu}^2\dd\ttheta<\infty.
\end{align}
For independent uniform samples $\ttheta_1,\ldots,\ttheta_N$ from $B$, put
\begin{align}
    \mu_{B,N}:=\frac{|B|}{N}\sum_{j=1}^N
      \gamma(\ttheta_j)\delta_{\ttheta_j},
    \qquad
    \mu_B:=\gamma\mathbf1_B\dd\ttheta.
\end{align}
Then
\begin{align}
    \EE\|S_\nu[\mu_{B,N}]-S_\nu[\mu_B]\|_{\H_\nu}^2
    \le\frac{|B|}{N}
    \int_B|\gamma(\ttheta)|^2
    \|\Phi_\ttheta\|_{\H_\nu}^2\dd\ttheta.
    \label{eq:truncated.mc.bound}
\end{align}
If $\mu_0=\gamma\dd\ttheta$ satisfies $S_\nu[\mu_0]=0$, then
\begin{align}
    \left(\EE\|S_\nu[\mu_{B,N}]\|_{\H_\nu}^2\right)^{1/2}
    &\le \|S_\nu[\mu_0-\mu_B]\|_{\H_\nu}+
    \left(\frac{|B|}{N}
    \int_B|\gamma(\ttheta)|^2
    \|\Phi_\ttheta\|_{\H_\nu}^2\dd\ttheta\right)^{1/2}.
    \label{eq:truncation.plus.sampling}
\end{align}
\end{lemma}

\begin{proof}
Apply the variance calculation in \eqref{eq:proof.finite.expected} to the
Hilbert-valued random variable
$|B|\gamma(\ttheta)\Phi_\ttheta$, and drop the nonnegative squared norm of
its mean.  This gives \eqref{eq:truncated.mc.bound}.  Minkowski's inequality
and $S_\nu[\mu_0]=0$ give \eqref{eq:truncation.plus.sampling}.
\end{proof}

\section{Details of the Numerical Illustration}\label{sec:num}
We use the one-dimensional target
\begin{align}
    f(x)=\sin(2\pi x)\mathbf 1_{[-1,1]}(x)
\end{align}
and the activation $\sigma(t)=\tanh t$.  Its Fourier transform, with the
convention of \cref{sec:pre}, is
\begin{align}
    \sigma^\sharp(\omega)
    =-\frac{i\pi}{\sinh(\pi\omega/2)}
\end{align}
in the sense of tempered distributions.

\subsection{Choice and Normalization of Ridgelet Functions}
Let $D(t)=\operatorname{dawsn}(t)$ denote the Dawson function.  Under our
Fourier convention,
\begin{align}
    D^\sharp(\omega)
    =-\frac{i\pi}{2}\sign(\omega)e^{-\omega^2/4}.
\end{align}
We use
\begin{align}
    \rho_1=D',\qquad
    \rho_2=c_2D'',\qquad
    \rho_3=D''',\qquad
    \rho_4=c_4D'''',
    \qquad
    \rho_{\rm nt}=\rho_2-\rho_4.
\end{align}
Parity immediately gives
$\iiprod{\sigma,\rho_1}=\iiprod{\sigma,\rho_3}=0$.  For the even derivative
orders, the two unnormalized pairings are
\begin{align}
    A_2
    &:=-\pi^2\int_0^\infty
      \frac{\omega e^{-\omega^2/4}}{\sinh(\pi\omega/2)}\dd\omega,
    & A_4
    &:=\pi^2\int_0^\infty
      \frac{\omega^3 e^{-\omega^2/4}}{\sinh(\pi\omega/2)}\dd\omega.
\end{align}
Adaptive one-dimensional quadrature gives
\begin{align}
    A_2&=-7.168527507803034,
    &A_4&=6.407939274274686,
\end{align}
and hence we set
\begin{align}
    c_2=A_2^{-1}=-0.13949866257909832,
    \qquad
    c_4=A_4^{-1}=0.15605641021203184.
\end{align}
Thus $\iiprod{\sigma,\rho_2}=\iiprod{\sigma,\rho_4}=1$ and
$\iiprod{\sigma,\rho_{\rm nt}}=0$.  Computing these constants from the
Fourier pairing is important: the earlier hand-tuned values $-1/1.5$ and
$1/4$ do not normalize the Dawson convention used by
\texttt{scipy.special.dawsn}.

\subsection{Deterministic Quadrature and Tail Stabilization}
All integrals in the displayed experiment are evaluated by tensor-product
Gauss--Legendre quadrature.  The spectra in the upper row of
\cref{fig:null.examples} are evaluated on a $200\times200$ display grid in
$[-6,6]^2$, using 256 Gauss--Legendre nodes for the $x$ integral, and each
spectrum is divided by its maximum absolute value for display.

For the reconstructions, write $Q_L[\rho]$ for the result obtained by
restricting $(a,b)$ to $[-L,L]^2$.  We use $L=40,80,160$.  At each $L$, the
$x$ integral uses
\begin{align}
    n_x(L)=\max\{160,4L\}
\end{align}
Gauss--Legendre nodes.  Each positive parameter half-axis uses $4L$ nodes.
The remaining quadrants are included algebraically using parity.  In
particular, $R[f;\rho_k](a,b)$ is even under
$(a,b)\mapsto(-a,-b)$ for $k=1,3$, whereas $\tanh(ax-b)$ is odd.  The paired
quadrature therefore cancels the $\rho_1$ and $\rho_3$ reconstructions
exactly.  For the odd ridgelet functions $\rho_2$ and $\rho_4$, we use that
the spectrum is odd in $a$ and even in $b$ to combine all four quadrants
without discarding any term.

Direct polynomial formulas for high Dawson derivatives lose significant
digits at large arguments.  For $|t|\ge8$, we therefore differentiate the
six-term asymptotic expansion
\begin{align}
    D(t)
    \sim\sum_{j\ge0}
    \frac{(2j-1)!!}{2^{j+1}t^{2j+1}}
\end{align}
term by term.  This stabilization is needed when $L$ is large.

The leading error remaining in $Q_L$ is finite-box truncation rather than
quadrature sampling.  We remove its first two inverse-power terms using the
three-level Richardson extrapolation
\begin{align}
    Q_\infty[\rho]
    \approx
    \frac13Q_{40}[\rho]-2Q_{80}[\rho]+\frac83Q_{160}[\rho].
\end{align}
The same nodes, weights, and extrapolation coefficients are used for
$\rho_2$ and $\rho_4$; consequently the displayed non-antipodal null result
is their numerical difference.

\subsection{Numerical Errors}
The error is measured at 1001 equally spaced evaluation points in
$[-1,1]$.  For admissible ridgelets it is measured from $f$, and for null
ridgelets from zero.
\begin{table}[tbp]
    \centering
    \caption{Errors of the five reconstructed panels in
    \cref{fig:null.examples}.}
    \label{tab:experiment.errors}
    \begin{tabular}{clrr}
        \toprule
        Panel & reference & RMSE & maximum absolute error\\
        \midrule
        (a) $\rho_1$ & $0$ & $0$ & $0$\\
        (b) $\rho_2$ & $f$ & $6.25\times10^{-4}$ & $1.01\times10^{-2}$\\
        (c) $\rho_3$ & $0$ & $0$ & $0$\\
        (d) $\rho_4$ & $f$ & $7.10\times10^{-4}$ & $5.95\times10^{-3}$\\
        (e) $\rho_{\rm nt}$ & $0$ & $1.09\times10^{-3}$ & $1.00\times10^{-2}$\\
        \bottomrule
    \end{tabular}
\end{table}
As a diagnostic, retaining the earlier coefficients and the small box
$[-5,5]^2$ while replacing Monte Carlo by converged Gauss--Legendre
quadrature still gives RMSE $0.2397$ and maximum residual $0.5317$ in panel
(e).  Hence the former sloping curve was not primarily a random-sampling
artifact: coefficient mismatch and finite-box bias were dominant.  The new
calculation reduces its RMSE by a factor of about 220.

The Monte Carlo estimate in \cref{lem:truncated.mc} continues to quantify
uniform random discretizations of a truncated null measure.  The figures
here instead use deterministic paired quadrature, so the sampling term in
\eqref{eq:truncation.plus.sampling} is replaced by high-order quadrature
error, while the Richardson extrapolation above controls the visible
finite-box bias.  The complete calculation and tabulated values are
reproduced by \path{20260714spectrum/plot_spectrum.py} in the accompanying
experiment repository.

\subsubsection*{Acknowledgements}
This work was supported by
JSPS KAKENHI 18K18113, 24K21316, 25H01453,
JST PRESTO JPMJPR2125,
JST BOOST JPMJBY24E2, and 
JST CREST JPMJCR2015, JPMJCR25I5.

The authors declare that they have no competing interests.

\bibliographystyle{abbrvnat}
\bibliography{summary_library} %

\end{document}